\documentclass[lettersize,journal]{IEEEtran}
\usepackage{amsmath,amsfonts}
\usepackage{algorithmic}
\usepackage{algorithm}
\usepackage{array}
\usepackage[caption=false,font=normalsize,labelfont=sf,textfont=sf]{subfig}
\usepackage{textcomp}
\usepackage{stfloats}
\usepackage{url}
\usepackage{verbatim}
\usepackage{graphicx}
\usepackage{cite}
\usepackage{xcolor}
\definecolor{rblue}{rgb}{0,0.5,1}
\definecolor{hollywoodcerise}{rgb}{0.96, 0.0, 0.63}
\definecolor{lasallegreen}{rgb}{0.03, 0.47, 0.19}
\definecolor{hanpurple}{rgb}{0.32, 0.09, 0.98}
\definecolor{green(pigment)}{rgb}{0.0, 0.65, 0.31}
\usepackage[pagebackref=false,breaklinks=true,colorlinks,bookmarks=false]{hyperref}
\hypersetup{colorlinks,linkcolor={red},citecolor={hanpurple},urlcolor={magenta}}  

\usepackage{subcaption}
\usepackage{amssymb}
\usepackage{multirow}
\usepackage{adjustbox}
\usepackage{booktabs} 

\usepackage[T1]{fontenc}
\usepackage[utf8]{inputenc} 
\usepackage{upquote}        
\usepackage{listings}

\lstdefinestyle{anyuserprompt}{
  basicstyle=\ttfamily\scriptsize,
  columns=fullflexible,
  breaklines=true,
  breakatwhitespace=false,
  keepspaces=true,
  showstringspaces=false,
  upquote=true,
  frame=single,
  rulecolor=\color{black},
  tabsize=2,
  xleftmargin=1ex,
  xrightmargin=1ex,
  numbers=left,
  numberstyle=\scriptsize,
  numbersep=6pt,
  postbreak=\mbox{\textcolor{gray}{$\hookrightarrow$}\space}
}

\begin{document}

\title{AnyUser: Translating Sketched User Intent into Domestic Robots}

\author{Songyuan Yang$^{*}$, Huibin Tan$^{*}$, Kailun Yang, Wenjing Yang, and Shaowu Yang$^{\dagger}$
\thanks{This work was supported by the Young Scientists Fund of the Hunan Natural Science Foundation (Grant No.2024JJ6474), the Youth Independent Innovation Science Fund Project of NUDT (Grant No.ZK24-08), the National Natural Science Foundation of China (Grant No. 62473139), the Hunan Provincial Research and Development Project (Grant No. 2025QK3019), the State Key Laboratory of Autonomous Intelligent Unmanned Systems (the opening project number ZZKF2025-2-10). \textit{(Songyuan Yang and Huibin Tan contributed equally to this work.) (Corresponding author: Shaowu Yang.)}}

\thanks{Songyuan Yang, Huibin Tan, Wenjing Yang, and Shaowu Yang are with the College of Computer Science and Technology, National University of Defense Technology, Changsha, Hunan 410073, China (e-mail: yangsongyuan@nudt.edu.cn; tanhb\_@nudt.edu.cn; wenjing.yang@nudt.edu.cn; shaowu.yang@nudt.edu.cn).}
\thanks{Kailun Yang is with the School of Artificial Intelligence and Robotics and the National Engineering Research Center of Robot Visual Perception and Control Technology, Hunan University, Changsha, Hunan 410082, China (e-mail: kailun.yang@hnu.edu.cn).}}

\markboth{IEEE Transactions on Robotics, April~2026}%
{Yang \MakeLowercase{\textit{et al.}}: AnyUser}

\maketitle

\begin{abstract}
We introduce AnyUser, a unified robotic instruction system for intuitive domestic task instruction via free-form sketches on camera images, optionally with language.  AnyUser interprets multimodal inputs (sketch, vision, language) as spatial-semantic primitives to generate executable robot actions requiring no prior maps or models. Novel components include multimodal fusion for understanding and a hierarchical policy for robust action generation. Efficacy is shown via extensive evaluations: (1) Quantitative benchmarks on the large-scale dataset showing high accuracy in interpreting diverse sketch-based commands across various simulated domestic scenes. (2) Real-world validation on two distinct robotic platforms, a statically mounted 7-DoF assistive arm (KUKA LBR iiwa) and a dual-arm mobile manipulator (Realman RMC-AIDAL), performing representative tasks like targeted wiping and area cleaning, confirming the system's ability to ground instructions and execute them reliably in physical environments. (3) A comprehensive user study involving diverse demographics (elderly, simulated non-verbal, low technical literacy) demonstrating significant improvements in usability and task specification efficiency, achieving high task completion rates (85.7\%-96.4\%) and user satisfaction. AnyUser bridges the gap between advanced robotic capabilities and the need for accessible non-expert interaction, laying the foundation for practical assistive robots adaptable to real-world human environments.
\end{abstract}


\section{Introduction}
\IEEEPARstart{T}{he} proliferation of service robotics in domestic environments has created unprecedented demand for intuitive human–robot interaction. Industrial settings benefit from structured workflows and deterministic commands; homes do not. Household deployments face unstructured scenes, diverse user demographics, and tasks that require flexible specification. Traditional instruction methods, whether programming by demonstration \cite{argall2009survey}, natural language interfaces \cite{tellex2011understanding, bisk2016natural, shridhar2020alfred}, or symbolic planning \cite{ghallab2004automated}, struggle to meet two concurrent requirements: (1) accessibility for non-experts across age groups and technical literacy levels, and (2) precise spatial grounding for reliable execution. The gap is most visible in long-horizon mobile manipulation, where coordinated spatial understanding and sequential planning are essential for tasks such as whole-house cleaning or multi-room delivery.

Existing approaches show three critical limitations. First, natural language interfaces \cite{tellex2011understanding, bisk2016natural, shridhar2020alfred} are intuitive for simple commands but ambiguous for complex spatial relations (e.g., “clean under the sofa while avoiding the power strip near the left leg”). Second, visual programming systems \cite{olsen2004fan, asenov2019vid2param} that depend on pre-defined maps or object databases adapt poorly to dynamic homes with frequent layout changes. Third, purely vision-based learning methods \cite{levine2018learning, florence2018dense, dasari2019robonet} often require environment-specific data at scale, limiting practical deployment. These factors create a persistent usability gap between robotic capability and everyday domestic needs, and this gap widens for elderly users or those less comfortable with technology.

\begin{figure*}[h]
    \centering
    \includegraphics[width=\linewidth]{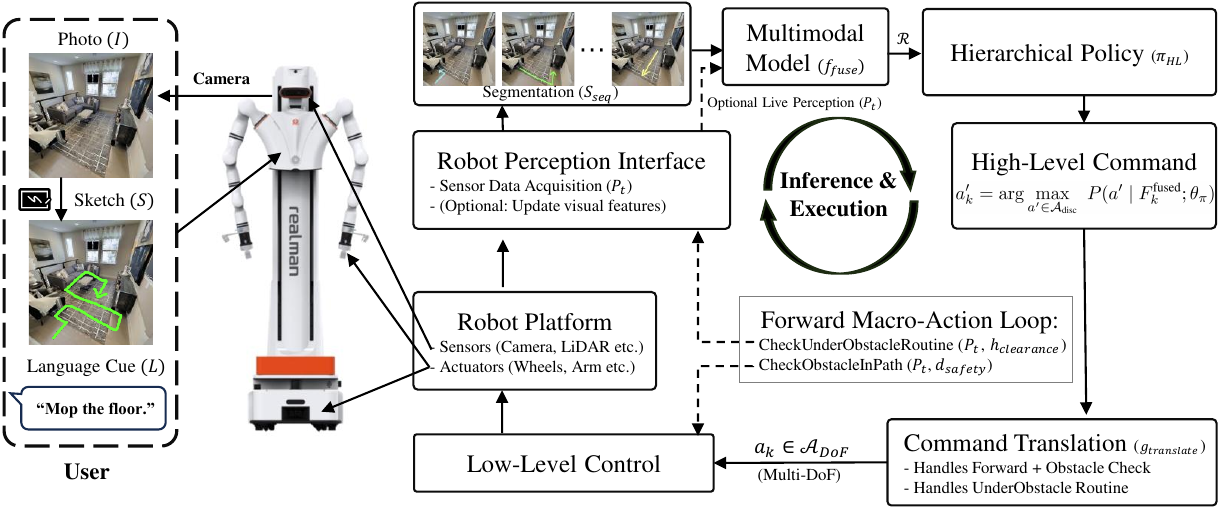}
    \caption{AnyUser architecture and runtime workflow. The user provides a third-person photograph $I$, draws sketches $S$ on the image, and may add a short language cue $L$. The sketch is deterministically segmented into an ordered sequence $S_{seq}$ and, together with $I$ and $L$, is encoded by the multimodal model $f_{fuse}$ to yield a runtime representation that conditions the hierarchical policy $\pi_{HL}$. For each segment the policy produces a high-level command $a'*k$ from the discrete set $A_{disc}$. During execution, egocentric live perception $P_t$ can be injected as an additional image-channel input for reactive checks such as obstacle presence and under-obstacle clearance. The command translation module $g_{translate}$ converts each high-level command into platform-specific multi-DoF control $a_k \in A_{DoF}$, which is executed by the robot’s low-level controllers in a closed loop.}
    \label{fig:inference}
    \vspace{-2mm}
\end{figure*}

We introduce AnyUser, a unified instruction system that combines photograph-grounded sketching with adaptive task reasoning. Users specify tasks by drawing free-form sketches directly on environment photographs and, if desired, adding brief language cues. As shown in Fig.~\ref{fig:inference}, this interaction lets novice users outline robot trajectories (e.g., vacuuming paths), operational zones (e.g., areas to avoid), and manipulation sequences (e.g., item pickup routes) through simple sketch-and-describe inputs. Unlike prior work that treats sketches primarily as low-level trajectories \cite{skubic2007using, turmukhambetov2015interactive, tanada2024sketch}, AnyUser interprets the multimodal tuple \((I,S,L)\) as a synergistic signal in which sketches provide spatial scaffolding, images supply visual-semantic grounding, and language refines intent.

\IEEEpubidadjcol

At the core of the system is an instruction understanding module that jointly processes geometric, visual-semantic, and linguistic streams. Modality-specific encoders extract salient features: a vision transformer for scene understanding, a geometric encoder that analyzes sketch structure via keypoints, and a transformer-based language encoder for contextual embeddings. A multimodal fusion network then associates sketch elements with visual regions using cross-modal attention and conditions interpretation on language. The result is a structured, spatially grounded runtime task representation \(\mathcal{R}\) that encodes targets, regions, and action semantics, bridging the gap between 2D interaction and 3D execution. The pipeline operates without pre-existing metric maps or object CAD models, relying on the user photograph and the robot’s online perception.

To translate \(\mathcal{R}\) into reliable behavior, AnyUser adopts a hierarchical action generation framework. A high-level policy, conditioned on \(\mathcal{R}\) and real-time perception \(P_t\), decomposes the task into platform-agnostic macro-actions (e.g., \texttt{forward}, \texttt{turn}, \texttt{check\_under}, \texttt{cover\_area}). An embodiment-specific translation module \(g_{translate}\) converts these macro-actions into low-level continuous control signals appropriate for the robot hardware (e.g., joint velocities or end-effector poses). Closed-loop perception enables reactive adjustments to obstacles and scene changes, supporting modularity, platform adaptability, and robust long-horizon execution.

This capability is supported by a hybrid training data strategy designed for generalization across heterogeneous homes. We build a composite dataset with more than 20{,}000 real indoor scenes from diverse households and 15{,}000 procedurally generated environments. The dataset, HouseholdSketch, combines the realism of real data with the scalability and targeted diversity of synthetic data, covering layouts, lighting, clutter, and sketch styles. This hybrid design reduces the gap between simulation and reality and enables robust performance in previously unseen settings.

The system targets three fundamental challenges in domestic robotics:

\textbf{Universal instruction accessibility.} Camera-based sketching replaces complex programming interfaces and lowers the entry barrier for non-experts. In user studies, 92\% of elderly participants (65+) specified cleaning tasks within three attempts, compared with 48\% using tablet-based GUIs.

\textbf{Cross-environment generalization.} Multimodal fusion of hand-drawn annotations, real-time perception, and language cues, together with hybrid training, allows adaptation to new layouts and object configurations without retraining.

\textbf{Long-horizon task reliability.} A hierarchical policy decomposes goals into macro-actions and executes them with closed-loop perception, resolving spatial ambiguities and maintaining progress on complex missions such as “clean all hardwood floors except under pet beds.”

Extensive experiments support the practicality of AnyUser. We evaluate on large-scale simulation benchmarks built on HouseholdSketch, on real deployments with two distinct platforms (a 7-DoF assistive arm and a dual-arm mobile manipulator), and in comprehensive user studies with diverse participants (\textit{N}=32). Results show reliable interpretation of sketch-based instructions, high task completion in simulation and on hardware, and clear usability gains for non-expert users, including older adults and participants with communication constraints. Our contributions are as follows:
\begin{itemize}
    \item A complete sketch-based instruction system for domestic robotics that integrates free-form visual annotations with real-time perception on photographic inputs, enabling intuitive task specification without pre-existing maps or object models.
    \item A multimodal instruction understanding framework that fuses geometric sketch structure, visual semantics, and optional language to produce a spatially grounded runtime task representation and executable macro-actions with high intent recognition across diverse user inputs.
    \item A hybrid training methodology that combines large-scale real-world data (20{,}000{+} indoor scenes) with procedurally generated synthetic environments (15{,}000), improving robustness and cross-environment generalization.
    \item Comprehensive validation in simulation and on hardware, including deployments on a 7-DoF assistive arm and a dual-arm mobile manipulator, demonstrating reliable execution of sketched tasks such as manipulation and area coverage.
    \item Rigorous user studies (\textit{N}=32) with diverse demographics showing clear usability gains for non-experts, including higher task completion and reduced instruction time compared with tablet-based GUI baselines.
    \item Development of HouseholdSketch, a large-scale dataset for sketch-based robot instruction with 35{,}000{+} annotated tasks, establishing a benchmark resource for visual–spatial human–robot interaction.
\end{itemize}

\section{Related Work}
\label{sec:related_work}

\subsection{Domestic Robotics}
\label{ssec:related_domestic}
The deployment of robots in domestic environments presents unique challenges compared to industrial settings, primarily due to environmental variability, the need for safety around humans, and the requirement for interaction with users lacking technical expertise \cite{kemp2007challenges, beer2012domesticated}. Early efforts focused on navigation and manipulation in semi-structured home scenarios, often relying on pre-existing maps or extensive environmental instrumentation \cite{pineau2003towards, saxena2008robotic}. While mobile cleaning robots (e.g., Roomba\cite{Roomba}, Neato\cite{Neato}) represent a commercial success, their capabilities are typically limited to 2D navigation on flat surfaces, often using relatively simple obstacle avoidance and area coverage algorithms \cite{koenig2004design}. More advanced domestic service robots, such as mobile manipulators designed for assistive tasks, require sophisticated perception, planning, and interaction capabilities \cite{srinivasa2010herb, chitta2012mobile}.

A critical aspect is enabling long-term autonomy and adaptation. Robots operating over extended periods must cope with changes in furniture layout, lighting conditions, and object configurations \cite{adkins2022probabilistic, kroemer2021review}. Research in lifelong learning and environment modeling aims to address this \cite{tipaldi2013lifelong, rusu2010semantic}, but often requires significant computational resources or expert oversight. Human-robot interaction (HRI) in the home context emphasizes naturalness and accessibility. While natural language interfaces (NLIs) offer intuitive command modalities \cite{tellex2011understanding, bisk2016natural, shridhar2020alfred}, they frequently suffer from spatial grounding ambiguities, making precise specification of geometric tasks (e.g., defining a complex cleaning path) challenging \cite{thomason2015learning, liang2023code}. Other interaction methods, like tangible interfaces or direct physical guidance \cite{argall2009survey}, may not scale well to complex, multi-step tasks or remote instruction. 

AnyUser targets this intersection. It provides a spatially precise yet accessible instruction modality that uses photograph-grounded sketching on \(I\) with optional language \(L\), operates without pre-existing maps, and adapts through real-time perception \(P_t\). The system interprets free-form sketches as spatial and semantic primitives fused with visual context to yield a runtime task representation \(\mathcal{R}\). This directly addresses the ambiguity of NLIs for geometric intent in path following and area definition, and improves accessibility for diverse users, including older adults \cite{pollack2002pearl}.

\subsection{Vision-based Learning for Robotics}
\label{ssec:related_vision}
Vision serves as a primary sensing modality for robots operating in unstructured environments, enabling tasks ranging from localization and mapping to object recognition and manipulation \cite{shahria2022comprehensive, robinson2023robotic}. Visual SLAM (Simultaneous Localization and Mapping) techniques allow robots to build maps and track their pose \cite{mur2015orb, engel2014lsd}, but often require careful initialization and can be sensitive to dynamic elements or textureless regions common in homes. Furthermore, traditional SLAM maps primarily capture geometry, lacking the semantic understanding needed for task-level reasoning \cite{chen2022overview}. Integrating semantic information (e.g., object labels, room categories) into maps has been an active research area \cite{salas2013slam++, mccormac2017semanticfusion}, but often relies on pre-trained object detectors or classifiers that may not generalize perfectly to novel home environments.

Recent advancements in deep learning have significantly impacted vision-based robotics. Convolutional Neural Networks (CNNs) \cite{lecun1998gradient} and Vision Transformers (ViTs) \cite{dosovitskiy2020image} excel at image-based tasks like semantic segmentation, depth estimation \cite{eigen2014depth}, and object detection \cite{redmon2016you, he2017mask}, providing rich perceptual information. This has fueled research in end-to-end learning, where policies directly map raw visual input to robot actions \cite{levine2016end, zhu2017target}, potentially simplifying the traditional perception–planning–control pipeline. However, end-to-end approaches often require large training sets, can be difficult to interpret, and may struggle to generalize across tasks or environments that differ from the training distribution \cite{pumacay2024colosseum, mysore2021regularizing}. Moreover, specifying complex and spatially precise tasks within an end-to-end framework remains challenging.

Vision-based instruction following, particularly linking natural language to visual scenes for robotic action \cite{anderson2018vision, lynch2020language}, has gained traction. These methods learn to ground linguistic commands in visual observations. AnyUser builds on these advances but adopts sketches as the primary spatial modality, complemented by the image \(I\) and optional language \(L\). The sketch acts as a strong geometric prior anchored in the photograph, reducing linguistic ambiguity for path specification and area definition. The system encodes the multimodal tuple \((I,S,L)\) with modality-specific backbones and fuses them to produce a spatially grounded runtime task representation \(\mathcal{R}\) and platform-agnostic macro-actions, without environment-specific fine-tuning. Using a static initial photograph for authoring and real-time perception \(P_t\) during execution balances instruction convenience with robustness to scene changes. Research on foundation models for robotics \cite{vemprala2024chatgpt, brohan2022rt} and large language models for complex instruction understanding \cite{ahn2022can, huang2023voxposer} continues to evolve; robust spatial grounding from diverse, non-expert inputs remains difficult. AnyUser addresses this by treating user-drawn sketches as first-class spatial–semantic primitives integrated tightly with vision and language.

\subsection{GUI-based Robot Programming Systems}
\label{ssec:related_gui}
Graphical User Interfaces (GUIs) offer an alternative to command-line or code-based programming, aiming to improve usability for non-programmers \cite{myers1990taxonomies}. In robotics, GUIs manifest in various forms. Visual Programming Languages (VPLs) like Blockly \cite{weintrop2017blockly} or Scratch \cite{Scratch} allow users to construct programs by connecting graphical blocks representing commands or control structures. While more intuitive than textual coding for simple sequences, VPLs can become cumbersome for complex tasks and often lack mechanisms for precise spatial specification relative to the real-world environment unless tightly integrated with simulation or pre-existing maps.

Map-based interfaces are prevalent, especially for mobile robots. Users often interact with a 2D map (either pre-scanned or built online) to specify navigation goals, draw paths, or define operational zones (e.g., keep-out areas for vacuum cleaners) \cite{biswas2012depth, thrun2002probabilistic}. These interfaces are effective when an accurate map exists and the environment remains largely static. However, they require an initial mapping phase, struggle with dynamic changes not reflected in the map, and may not easily support tasks requiring interaction with specific objects or 3D spatial reasoning (e.g., ``clean under the chair''). Furthermore, the abstraction level of a 2D map can sometimes make it difficult for users to correlate with the physical 3D space.

Programming by Demonstration (PbD) systems allow users to teach robots tasks by physically guiding them or using teleoperation interfaces, often supplemented by GUIs for refining or segmenting the demonstration \cite{argall2009surveydemonstration, billard2008survey}. While intuitive for kinesthetic teaching, PbD can be time-consuming, requires the user to be physically present (or use complex teleoperation setups), and generalization from demonstrations remains a challenge \cite{osa2018algorithmic}.

Sketch-based interfaces represent a related but distinct category. Early work explored sketches for path specification or simple behavior generation \cite{skubic2003sketch, skubic2004sketch}. These systems often treated sketches primarily as geometric trajectories or required calibrated setups. More recent work has explored learning from sketches for trajectory generation or task definition \cite{zhi2024instructing, yu2025sketch}, sometimes combining sketches with other modalities like language \cite{sundaresan2024rt, gu2023rt}. However, many existing sketch-based systems process the sketch in isolation or assume a static, fully observable environment.

AnyUser advances this line by: (1) interpreting free-form sketches drawn on real photographs as spatial–semantic primitives rather than only low-level trajectories; (2) fusing the multimodal tuple \((I,S,L)\) with modality-specific encoders to produce a spatially grounded runtime task representation \(\mathcal{R}\); (3) operating without pre-existing maps or environment models by relying on the authoring photograph and real-time perception \(P_t\) at execution; and (4) demonstrating robust performance on long-horizon tasks in diverse, previously unseen domestic environments. By grounding instructions directly in the user’s view of the scene, the approach avoids the abstraction inherent to typical GUIs or VPLs and provides an accessible interaction paradigm for spatially complex tasks. Related explorations in augmented reality interfaces \cite{smith2024augmented} share the goal of intuitive spatial specification; the photograph-based workflow in AnyUser offers a lightweight alternative in terms of hardware and authoring effort.

\section{AnyUser System Design} \label{sec:system_design}
This section presents the design principles of AnyUser. We first formalize the problem, then describe the learning paradigm for interpreting multimodal instructions and generating robot actions, and finally outline the data strategy chosen for deployment in diverse homes.

\begin{table}[h]
\centering
\caption{Main notations used in our paper.}
\label{tab:notation}
\resizebox{\columnwidth}{!}{%
\begin{tabular}{ll}
\hline
Symbol & Meaning \\
\hline
\(I,\,S,\,L\) & Environment photo, user sketches, language cue \\
\(\mathcal{I}=(I,S,L)\) & Multimodal instruction tuple \\
\(\mathcal{S}_{\text{seq}}=\{s_k\}_{k=1}^{N_{\text{seg}}}\) & Ordered post-segmentation sketch segments; \(N_{\text{seg}}\) is count \\
\(x_t,\,P_t\) & Robot state and live perception at time \(t\) \\
\(\phi_V,\,\phi_S,\,\phi_L\) & Visual, sketch, and language encoders \\
\(\psi_{\text{fuse}}\) & Multimodal fusion and spatial grounding module \\
\(f_{\text{fuse}}\) & Instruction-understanding network mapping \((I,S,L)\!\rightarrow\!\mathcal{R}\) \\
\(\mathcal{R},\,\mathcal{R}_{\text{ann}}\) & Runtime task representation; annotation ground truth \\
\(\mathcal{M}\) & Robot platform \\
\(\mathcal{G}\) & Latent user goal/intent \\
\(\pi_{\text{HL}}\) & High-level policy \\
\(g_{\text{translate}}\) & Command translator \\
\(\mathcal{A},\,\mathcal{A}_{\text{disc}}\) & Abstract macro-action space; discrete macro-action set \\
\(a'_k,\,a_k\) & Predicted macro-action; translated low-level command \\
\(\mathcal{A}_{\text{DoF}}\) & Platform-specific continuous control space \\
\hline
\end{tabular}}
\end{table}

\subsection{Problem Definition}
The goal is to translate intuitive, human-generated multimodal instructions into reliable, spatially grounded actions in unstructured domestic settings. Let the user provide \(\mathcal{I}=(I,S,L)\), where \(I\in\mathbb{R}^{H\times W\times 3}\) is a photograph of the target scene, \(S=\{l_1,\dots,l_n\}\) is a set of free-form trajectories drawn on \(I\) with each \(l_i\subset\mathbb{R}^2\), and \(L\) is an optional language cue. The robot, characterized by platform \(\mathcal{M}\), operates with state \(x_t\in X\) and real-time perception \(P_t\in\mathcal{P}\) (e.g., RGB–D, proprioception). 

The system infers a spatially grounded runtime representation \(\mathcal{R}\) from \(\mathcal{I}\) and uses a policy \(\pi\) to generate an action sequence \(A=\{a_1,\dots,a_T\}\) that achieves the user’s latent goal \(\mathcal{G}\) in environment \(E\). Formally,
\[
\pi:\; X \times \mathcal{P} \times \mathcal{R} \rightarrow \mathcal{A},
\]
where \(\mathcal{A}\) denotes the abstract action space. The mapping must bridge the gap between the two-dimensional authoring interface \((I,S,L)\) and three-dimensional execution, requiring: (i) robust spatial grounding of sketches in the scene, (ii) disambiguation of uncertainties in sketches and language, and (iii) adaptation to changes not captured in the initial photograph. The design intentionally avoids reliance on pre-existing metric maps or detailed object models; environmental understanding arises from the user photograph and online perception. For training and quality assurance, we use \(\mathcal{R}_{\text{ann}}\) as annotation ground truth, which does not appear at runtime. 

\begin{figure*}[h]
    \centering
    \includegraphics[width=\linewidth]{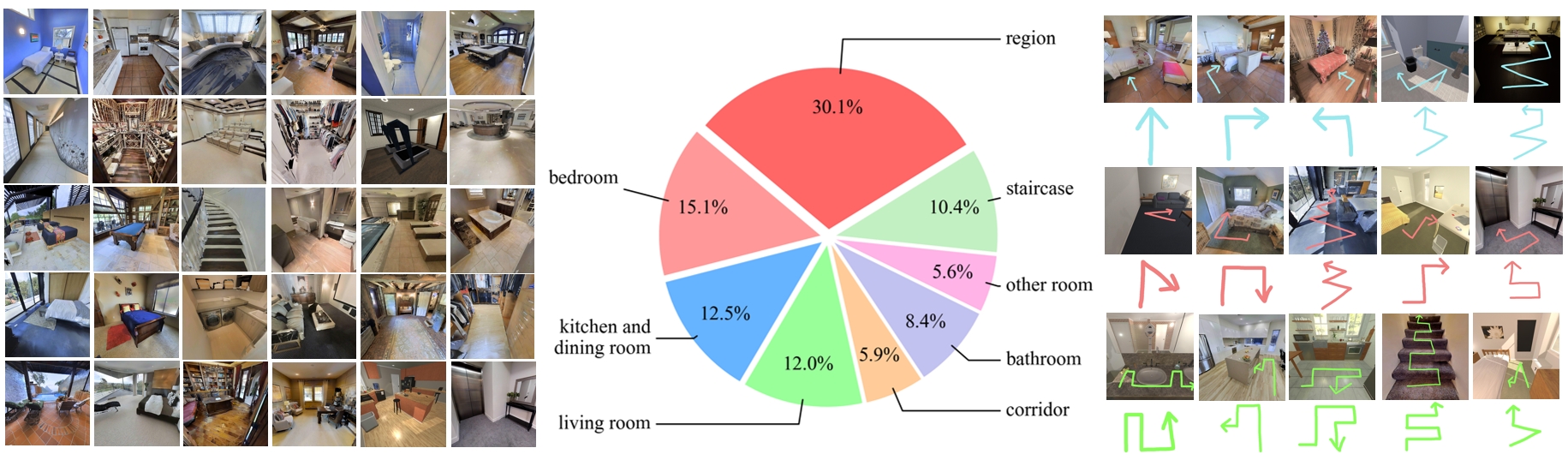}
    \caption{Overview of the HouseholdSketch dataset utilized for training and evaluation. Left: A selection of representative images illustrating the visual diversity of indoor environments included. Center: Pie chart depicting the proportional distribution of scene categories within the dataset. Right: Exemplar sketch inputs overlaid on corresponding scene images and shown in isolation.}
    \label{fig:dataset}
\end{figure*}

\subsection{Multimodal Instruction Learning}
A central tenet of AnyUser is to treat the user input \(\mathcal{I}=(I,S,L)\) as a synergistic multimodal signal rather than independent channels or mere geometric directives. The sketch set \(S\) provides the primary spatial scaffolding by delineating trajectories, regions of interest, and object references in the context of the photograph \(I\). Free-form sketches are expressive but imprecise, so they require contextualization. The image \(I\) supplies visual–semantic evidence that links sketched elements to scene structure (e.g., floors, walls, furniture), to objects of interest (e.g., obstacles to avoid, items to manipulate), and to traversable space. Optional language \(L\) complements these cues by refining intent, adding constraints (e.g., “vacuum under the table but not the rug”), and disambiguating actions (e.g., “wipe this countertop”).

To exploit this input, AnyUser uses an instruction understanding module \(f_{\text{fuse}}\) that jointly processes the geometric (\(S\)), visual–semantic (\(I\)), and linguistic (\(L\)) streams. The module maps \(\mathcal{I}\) to a structured, spatially grounded runtime representation \(\mathcal{R}\), i.e., \(\mathcal{R}=f_{\text{fuse}}(I,S,L)\). Modality-specific encoders first extract features: a visual backbone \(\phi_V\) produces dense spatial and semantic features \(F_V=\phi_V(I)\); a sketch encoder \(\phi_S\) computes sketch features \(F_S=\phi_S(S)\); and a transformer-based language encoder \(\phi_L\) yields contextual embeddings \(F_L=\phi_L(L)\). These features are later fused to produce \(\mathcal{R}\) used by the policy.

The unimodal features \((F_V, F_S, F_L)\) are integrated by a multimodal fusion network \(\psi_{\text{fuse}}\). Cross-modal attention lets sketch features in \(F_S\) associate with relevant visual regions in \(F_V\), while language embeddings in \(F_L\) modulate both channels. The fused output is the runtime task representation \(\mathcal{R}\). This representation encodes the inferred intent, including spatial parameters such as target waypoints, regions of interest or avoidance grounded in \(I\), and functional semantics such as task type and constraints derived from \(L\) and visual context. \(\mathcal{R}\) provides the structured input that conditions the high-level policy \(\pi_{\text{HL}}\). During training, \(f_{\text{fuse}}=\psi_{\text{fuse}}\circ(\phi_V,\phi_S,\phi_L)\) is optimized so that \(\mathcal{R}\) aligns with the annotation ground truth \(\mathcal{R}_{\text{ann}}\) and supports successful execution toward the latent goal \(\mathcal{G}\).

\subsection{Hierarchical Action Generation for Multi-DoF Control}
Translating the spatially grounded representation \(\mathcal{R}\) into physical behavior requires action sequences that are temporally coherent and consistent with robot dynamics. Target domestic tasks such as vacuuming along a sketched path, wiping a designated area, or navigating for object retrieval often demand precise multi-DoF control in cluttered spaces. Manipulation further increases the degrees of freedom and the need for closed-loop adjustments.

To accommodate diverse robot embodiments \(\mathcal{M}\), AnyUser adopts a hierarchical action generation framework. A high-level policy \(\pi_{\text{HL}}\) conditions on the robot state \(x_t\), real-time perception \(P_t\), and \(\mathcal{R}\) to produce platform-agnostic macro-actions:
\[
\pi_{\text{HL}}:\; X \times \mathcal{P} \times \mathcal{R} \rightarrow \mathcal{A},
\]
where \(\mathcal{A}\) denotes an abstract macro-action set. This abstraction supports modularity and robustness by separating task-level decisions from embodiment-specific control. 

An embodiment-specific translation module \(g_{\text{translate}}\) converts macro-actions produced by \(\pi_{\text{HL}}\) into low-level commands for the target platform. Formally,
\[
g_{\text{translate}}:\; \mathcal{A} \times \mathcal{M} \rightarrow \mathcal{A}_{\text{DoF}},
\]
where \(\mathcal{A}\) is the abstract macro-action set and \(\mathcal{A}_{\text{DoF}}\) denotes the continuous, potentially multi-DoF control space of the robot. For a differential-drive base, \(a_t \in \mathcal{A}_{\text{DoF}}\) may be a velocity pair \((v_t,\omega_t)\in\mathbb{R}^2\). For a \(k\)-DoF manipulator, \(a_t\) may be joint velocities \(\dot{q}_t\in\mathbb{R}^k\) or an end-effector twist \(\dot{x}_{ee}\in\mathbb{R}^6\). The translator encapsulates kinematics, platform interfaces, and low-level controllers. Execution is closed-loop: \(\pi_{\text{HL}}\) and \(g_{\text{translate}}\) leverage perception \(P_t\) not only for state estimation but also for reactive adjustment of macro-action parameters and sequencing to handle obstacles, execution errors, and discrepancies between the authoring photograph \(I\) and the current scene, enabling robust long-horizon behavior.

\subsection{Hybrid Training Data Strategy for Domestic Environments}
The generalization and real-world performance of AnyUser, especially the perception module \(f_{\text{fuse}}\) and the high-level policy \(\pi_{\text{HL}}\), depend on the diversity and scale of training data. Domestic environments vary widely in layout, lighting, object types and arrangements, clutter, textures, and human activity. Capturing this spectrum is essential for robust deployment. Relying only on synthetic data offers scalability and convenient ground-truth generation, but the mismatch between simulation and reality can degrade performance due to differences in sensor noise, lighting realism, material properties, and object appearance. Using only real data provides high fidelity, yet large-scale collection and annotation are costly and labor-intensive, and coverage of safety-critical edge cases or rare configurations may remain limited.

To mitigate these limitations, AnyUser adopts a hybrid strategy with a composite dataset \(\mathcal{D}=\mathcal{D}_{\text{real}}\cup\mathcal{D}_{\text{synth}}\). The real component \(\mathcal{D}_{\text{real}}\) contains data collected in diverse households. Each sample \(d_{\text{real}}\in\mathcal{D}_{\text{real}}\) typically includes a tuple \((I_j,S_j,L_j)\) captured in situ, optionally paired with an annotation ground truth \(\mathcal{R}_{\text{ann},j}\) that reflects the intended specification, or with a successful execution trace \(A^{\star}_j=\{a_{j,1},\dots,a_{j,T_j}\}\) obtained from expert demonstrations or post-hoc recovery. The synthetic component \(\mathcal{D}_{\text{synth}}\) consists of procedurally generated domestic scenes curated to complement \(\mathcal{D}_{\text{real}}\). These samples target challenging navigation layouts, varied furniture styles, multi-step manipulation patterns, diverse sketch styles, and difficult perceptual conditions such as extreme illumination, occlusions, and specific textures. The hybrid design combines the ecological validity of \(\mathcal{D}_{\text{real}}\) with the scalability and controllability of \(\mathcal{D}_{\text{synth}}\), improving robustness, generalization, and coverage of rare yet important cases.

\begin{figure*}[h]
    \centering
    \includegraphics[width=\linewidth]{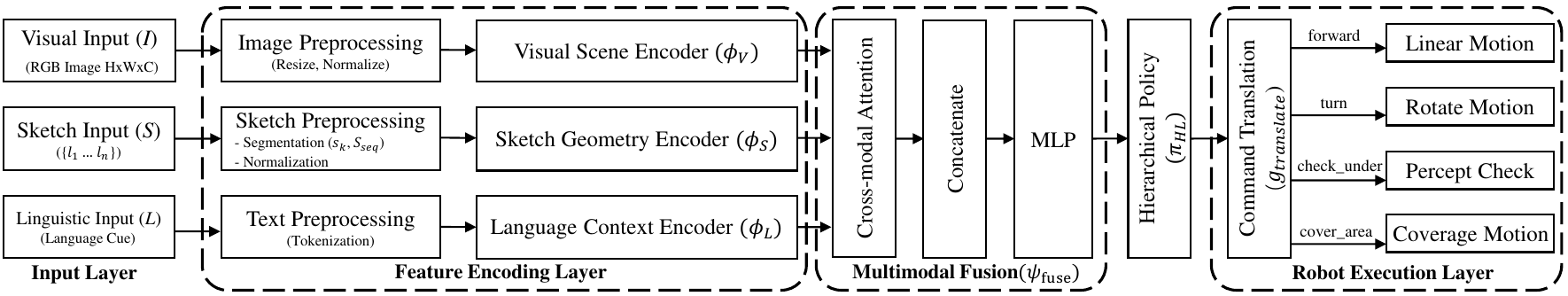}
    \caption{Detailed pipeline of the AnyUser's architecture. The Input Layer receives Visual ($I$), Sketch ($S$), and Linguistic ($L$). After preprocessing, encoders extract modality features. Fusion $\psi_{\text{fuse}}$ aligns sketch with image via cross-modal attention and an MLP, and the hierarchical policy \(\pi_{\text{HL}}\) predicts a macro-action per segment. Command translation $g_{\text{translate}}$ converts these macros into platform-specific multi-DoF primitives for execution.}
    \label{fig:pipeline}
\end{figure*}

\section{Methods} \label{sec:methods}
Building on Sec.~\ref{sec:system_design}, this section specifies the concrete realization of AnyUser. We first detail the construction of the HouseholdSketch dataset. We then describe the multimodal network that maps \(\mathcal{I}=(I,S,L)\) to the runtime representation \(\mathcal{R}\). Next, we instantiate the high-level policy \(\pi_{\text{HL}}\) as a compact discrete macro-action controller over \(\mathcal{A}_{\text{disc}}\) and present Algorithm~\ref{alg:inference_loop_revised}. We conclude with the optimization objectives and training procedure, followed by the runtime inference and the platform translation module \(g_{\text{translate}}\) used for deployment.

\subsection{Dataset Construction}
To advance research in sketch-based human–robot interaction for domestic applications, we introduce HouseholdSketch (Fig.~\ref{fig:dataset}), a large-scale hybrid dataset engineered to capture the complexity of real homes while providing precise geometric and semantic supervision. The corpus integrates procedurally generated synthetic scenes with meticulously curated real-world observations under a unified annotation protocol and a multi-stage validation workflow. The goal is to supply training and evaluation data that reflect the diversity of domestic settings and that support deployment-oriented instruction grounding.

\textbf{Synthetic pipeline.} We leverage four industry-standard simulation platforms, Gibson \cite{li2021igibson}, AI2-THOR \cite{kolve2022ai2thor}, Matterport3D \cite{matterport3d}, and VirtualHome \cite{8578984}, to generate 15{,}000 photorealistic scene viewpoints \( (I\in\mathbb{R}^{H\times W\times 3}) \) under controlled parametric variation. Each viewpoint samples illumination spectra and intensity, camera extrinsics, furniture topology, and object configuration, yielding broad coverage of geometric layouts and material reflectance properties. This design allows targeted stress testing of corner cases such as narrow passages, reflective surfaces, and cluttered floor regions that are difficult to curate at scale in the real world.

\textbf{Real-world corpus.} To reduce the gap between simulation and reality, we complement the synthetic set with 20{,}000 scene captures from 50 distinct residences spanning diverse architectural styles (open plan and multi-room), interior conditions (from pristine to highly cluttered), and illumination regimes (daylight and artificial lighting). Data are collected with calibrated multi-sensor rigs to ensure metrological consistency and to preserve high-fidelity texture, complex inter-reflections, and naturally occurring object arrangements that are challenging to reproduce in simulation.

\textbf{Unified annotation schema.} For each scene \(I\), trained annotators use a custom web interface to create multimodal instruction tuples \((S,L)\). The sketch set \(S=\{l_1,\dots,l_n\}\), \(l_i\subset\mathbb{R}^2\), encodes task geometry via free-form strokes drawn directly on the photograph. The language cue \(L\) provides concise constraints that disambiguate intent (e.g., “wipe countertop avoiding plant”, “confirm reachability under cabinet”). We cover three instruction classes systematically: \textit{path specification} (navigation or manipulation trajectories), \textit{area definition} (regions for cleaning or avoidance), and \textit{interaction queries} (spatial reasoning tasks). Each \((I,S,L)\) tuple is paired with annotation ground truth \(\mathcal{R}_{\text{ann}}\). For synthetic scenes, \(\mathcal{R}_{\text{ann}}\) is derived from simulator state, yielding millimeter-accurate 3D waypoints and object meshes. For real scenes, \(\mathcal{R}_{\text{ann}}\) is obtained through photogrammetric reconstruction and robotic execution traces, capturing operational constraints with \(\pm 2\)~cm positional fidelity.

\textbf{Quality assurance.} Quality control is embedded throughout the lifecycle via a three-tier validation cascade. First, annotators pass competency tests using a reference set of 500 pre-validated samples. Second, a dedicated QA team performs cross-modal consistency checks that verify geometric alignment between \(S\) and \(\mathcal{R}_{\text{ann}}\), assess the relevance of \(L\) to both \(I\) and \(S\), and confirm physical plausibility under scene constraints. Third, a subset of 1{,}200 samples undergoes empirical validation through robotic execution trials to confirm operational realizability. This process rejects 18.7\% of initial annotations during iterative refinement, resulting in 35{,}000 annotated task instances with strong cross-modal consistency and deployability. The final corpus exhibits broad scene and task diversity, spanning 12 domestic activity categories and 43 fine-grained object classes, and serves as a high-value benchmark for instruction grounding in household robotics.

\subsection{Model Details}

The core computational component of AnyUser is a multimodal instruction interpretation network that maps the heterogeneous input \(\mathcal{I}=(I,S,L)\) to a spatially grounded runtime representation \(\mathcal{R}\), which conditions a high-level policy \(\pi_{\text{HL}}\) to produce a sequence of platform-agnostic macro-actions \(A'=\{a'_k\}_{k=1}^{N_{\text{seg}}}\subset\mathcal{A}_{\text{disc}}\). This requires geometric understanding of free-form sketches, semantic grounding in the visual scene, and contextual refinement from language. The architecture, illustrated in Fig.~\ref{fig:pipeline}, follows a staged pipeline: input parametrization, modality-specific encoding, cross-modal fusion to form \(\mathcal{R}\), and macro-action reasoning with \(\pi_{\text{HL}}\).

\subsubsection{Input Parametrization and Preprocessing}
\label{sssec:input_repr}
The system consumes three modalities: the visual scene, user-drawn sketches, and optional language. Each modality is preprocessed into a representation compatible with the downstream encoders and the fusion module.

\textbf{Visual input (\(I\)).} The RGB image \(I\in\mathbb{R}^{H\times W\times 3}\) that captures the operational environment is resized to \(224\times224\) and normalized with ImageNet mean and variance \cite{deng2009imagenet}, matching the requirements of the visual encoder \(\phi_V\).

\textbf{Sketch input (\(S\)).} A sketch \(S\) is a set of trajectories \(\{l_1,\dots,l_n\}\). Each trajectory \(l_i\) is represented as a sequence of \(k_i\) pixel coordinates \(l_i=\{p_{i,1},\dots,p_{i,k_i}\}\) with \(p\in[0,H-1]\times[0,W-1]\). To bridge free-form strokes with the discrete macro-action space \(\mathcal{A}_{\text{disc}}\), all trajectories are deterministically segmented into geometric primitives and concatenated into a single ordered sequence
\begin{equation}
\mathcal{S}_{\text{seq}}=\{s_{1},s_{2},\dots,s_{N_{\text{seg}}}\}.
\end{equation}
A new segment boundary is created when either (i) the turning angle at a point triplet \((p_{t-1},p_t,p_{t+1})\) exceeds the threshold \(\theta_{\text{turn}}\), or (ii) the Euclidean length of the current segment exceeds \(L_{\text{max}}\). This produces segments \(s_{k}=\{p_{k,1},\dots,p_{k,m_{k}}\}\). For numerical stability, coordinates within each segment are normalized to \([-1,1]^2\). Each segment \(s_{k}\) represents an atomic motion intention and later conditions the macro-action predictor. The total number of segments over \(S\) is denoted \(N_{\text{seg}}\).

\textbf{Language input (\(L\)).} The optional natural language cue \(L\) is tokenized into subword units compatible with the language encoder \(\phi_L\), providing concise semantic constraints that refine intent and disambiguate sketch semantics.

\subsubsection{Multimodal Feature Extraction}
\label{sssec:modality_encoding}
Each preprocessed modality is encoded by a dedicated neural module into a fixed-dimensional representation that is later fused to form \(\mathcal{R}\).

\textbf{Sketch geometry encoder (\(\phi_S\)).} The goal of \(\phi_S\) is to extract a geometrically salient feature \(f^{S}_{k}\in\mathbb{R}^{D_S}\) for each segment \(s_{k}\). We adopt a structured analysis centered on keypoint identification. A lightweight detector \(\phi_{\text{KP}}\) locates structurally significant points in \(s_{k}\), typically the start, the end, and high-curvature corners. Let \(K_{k}=\{\kappa_{k,1},\dots,\kappa_{k,N_\kappa}\}\) denote the detected keypoints, where each \(\kappa\) has normalized coordinates \(\kappa_{\text{loc}}\in[-1,1]^2\) and a type \(\kappa_{\text{type}}\in\{\text{Start},\text{End},\text{Corner}\}\). The detector applies 1D convolutions along the ordered point sequence of \(s_{k}\), followed by attention or pooling to score candidate keypoints. Per-keypoint embeddings are then computed and aggregated with an MLP to summarize the segment’s geometry. We set \(D_S=256\).
\begin{equation}
K_{k}=\phi_{\text{KP}}(s_{k}),
\end{equation}
\begin{equation}
f^{S}_{k}=\mathrm{Aggregate}_{\theta_{\text{agg}}}\big(\{\mathrm{encode}(\kappa)\mid \kappa\in K_{k}\}\big),
\end{equation}
where \(\mathrm{encode}(\kappa)\) maps a keypoint to a feature vector and \(\mathrm{Aggregate}_{\theta_{\text{agg}}}\) denotes the aggregation function with parameters \(\theta_{\text{agg}}\). The resulting \(f^{S}_{k}\) conditions downstream fusion.

\textbf{Visual scene encoder (\(\phi_V\)).} We use a Vision Transformer \cite{vit} as \(\phi_V\). Given the normalized image \(I\), the network applies patch embedding followed by transformer layers. We extract two outputs: a class-token embedding \(f^{V}_{\text{cls}}\in\mathbb{R}^{D_V}\) that summarizes the scene, and a grid of patch features \(F^{V}_{\text{grid}}\in\mathbb{R}^{N_p\times D_V}\) that preserves spatial detail, where \(N_p=196\). The attention structure of ViT supports cross-modal association in the fusion stage. During training, \(\phi_V\) is frozen to leverage its pre-trained visual understanding. We set \(D_V=768\).

\textbf{Language context encoder (\(\phi_L\)).} We adopt the text transformer from CLIP \cite{clip} as \(\phi_L\). The tokenized sequence for \(L\) is mapped to contextual embeddings, from which we use the [CLS]-pooled sentence vector \(f^{L}\in\mathbb{R}^{D_L}\). The language encoder is kept frozen during training. We set \(D_L=512\).

\subsubsection{Multimodal Feature Fusion and Grounding (\(\psi_{\text{fuse}}\))}
\label{sssec:fusion}
For each sketch segment \(s_{k}\), the fusion module \(\psi_{\text{fuse}}\) integrates unimodal features to produce a representation that captures intent, visual evidence, and linguistic context. Spatial grounding is achieved with cross-modal attention, where the sketch feature \(f^{S}_{k}\) queries the visual patch features \(F^{V}_{\text{grid}}\). The attention weights \(\alpha_{k,p}\) measure the relevance of patch \(p\) to segment \(s_{k}\):
\begin{equation}
e_{k,p}=\frac{(Q f^{S}_{k})^{\top}(K F^{V}_{\text{grid},p})}{\sqrt{d_{\text{key}}}},\qquad
\alpha_{k,p}=\frac{\exp(e_{k,p})}{\sum_{p'=1}^{N_p}\exp(e_{k,p'})},
\end{equation}
where \(Q\) and \(K\) are learnable projections and \(d_{\text{key}}\) is the key dimension. The attended visual feature is the value-weighted sum
\begin{equation}
f^{V}_{\text{att},k}=\sum_{p=1}^{N_p}\alpha_{k,p}\big(V F^{V}_{\text{grid},p}\big),
\end{equation}
with \(V\) a learnable projection. We then concatenate the relevant features and project them with an MLP to obtain the fused segment representation
\begin{equation}
F^{\text{fused}}_{k}=\mathrm{MLP}_{\text{fuse}}\big([\,f^{S}_{k};\, f^{V}_{\text{att},k};\, f^{V}_{\text{cls}};\, f^{L}\,]\big),
\label{eq:fusion_revised}
\end{equation}
where \([\,;\,]\) denotes concatenation. We set the fused dimension \(D_{\text{fused}}=512\). The collection \(\{F^{\text{fused}}_{k}\}_{k=1}^{N_{\text{seg}}}\) over all segments forms the structured input used to construct the runtime representation \(\mathcal{R}\) that conditions \(\pi_{\text{HL}}\).

\subsubsection{Hierarchical Policy for Task Decomposition and Command Generation (\(\pi_{\text{HL}}\))}
\label{sssec:reasoning}
Given the fused segment representations \(\{F^{\text{fused}}_{k}\}_{k=1}^{N_{\text{seg}}}\), the high-level policy \(\pi_{\text{HL}}\) decomposes the sketched instruction into a sequence of discrete macro-actions. For each segment representation \(F^{\text{fused}}_{k}\), \(\pi_{\text{HL}}\) predicts one action \(a'_{k}\) from a compact vocabulary implemented as an MLP classifier:
\begin{equation}
\pi_{\text{HL}}:\ \mathbb{R}^{D_{\text{fused}}}\ \rightarrow\ \mathbb{R}^{|\mathcal{A}_{\text{disc}}|},\quad
z_{k}=\pi_{\text{HL}}\!\left(F^{\text{fused}}_{k}\right),
\end{equation}
where \(z_{k}\) are the logits over the predefined action set
\begin{align}
\mathcal{A}_{\text{disc}}
= \{\ &\texttt{forward},\ \texttt{turn\_p45},\ \texttt{turn\_n45},\notag\\
      &\texttt{turn\_p90},\ \texttt{turn\_n90},\notag\\ &\texttt{check\_under},\ \texttt{cover\_area}\ \}.
\label{eq:action_space}
\end{align}
The class probabilities are obtained with a softmax:
\begin{equation}
P(a'=c \mid F^{\text{fused}}_{k};\theta_{\pi})
= \frac{\exp(z_{k,c})}{\sum_{c'\in\mathcal{A}_{\text{disc}}}\exp(z_{k,c'})},
\end{equation}
and the predicted action is
\begin{equation}
a'_{k}=\arg\max_{a'\in\mathcal{A}_{\text{disc}}}\ P(a'\mid F^{\text{fused}}_{k};\theta_{\pi}).
\label{eq:command_prediction_final}
\end{equation}
Applying this procedure over all segments yields the macro-action sequence
\(A'=\{a'_{k}\}_{k=1}^{N_{\text{seg}}}\).
When a segment corresponds to an area-level token, selecting \texttt{cover\_area} triggers serpentine coverage as specified in Algorithm~\ref{alg:inference_loop_revised}.

\subsubsection{Optimization Objectives}
\label{sssec:training_obj}
The trainable parameters of the sketch encoder \(\phi_{S}\), the fusion module \(\psi_{\text{fuse}}\), and the high-level policy \(\pi_{\text{HL}}\) (with \(\phi_{V}\) and \(\phi_{L}\) frozen) are optimized by minimizing a composite objective \(\mathcal{L}_{\text{total}}\). The loss aggregates supervision signals for action prediction accuracy, geometric keypoint quality, and trajectory alignment.

\textbf{Action command prediction loss (\(\mathcal{L}_{\text{task}}\)).}
For each segment \(s_{k}\) the model predicts a macro-action \(a'_{k}\). Let \(a'^{\star}_{k}\) denote the ground-truth class. The cross-entropy over all segments in a mini-batch is
\begin{equation}
\mathcal{L}_{\text{task}}
= - \frac{1}{N_{\text{seg}}^{\text{tot}}}
\sum_{(b,k)\in\mathcal{S}_{\text{batch}}}
\log P\!\left(a'^{\star}_{k}\ \middle|\ F^{\text{fused}}_{k};\,\theta_{\pi}\right),
\label{eq:task_loss_final}
\end{equation}
where \(\mathcal{S}_{\text{batch}}\) indexes all segments in the mini-batch, \(N_{\text{seg}}^{\text{tot}}=|\mathcal{S}_{\text{batch}}|\), \(F^{\text{fused}}_{k}\) is the fused feature of segment \(s_{k}\), and \(\theta_{\pi}\) are the parameters of \(\pi_{\text{HL}}\).

\textbf{Sketch keypoint supervision loss (\(\mathcal{L}_{\text{kp}}\)).}
To guide the sketch encoder \(\phi_{S}\) toward geometrically salient structure, we supervise the keypoint detector \(\phi_{\text{KP}}\). For each segment \(s_{k}\), let \(K^{\star}_{k}\) be the set of ground-truth keypoints with location \(\boldsymbol{\kappa}_{\text{loc}}\in[-1,1]^2\) and type \(\kappa_{\text{type}}\in\{\text{Start},\text{End},\text{Corner}\}\). The detector produces, at the matched positions, a location estimate \(\hat{\boldsymbol{\kappa}}_{\text{loc}}(\kappa)\) and a type probability vector \(\hat{\mathbf{p}}_{\text{type}}(\kappa)\) for each \(\kappa\in K^{\star}_{k}\).
The keypoint loss over a mini-batch is
\begin{align}
\mathcal{L}_{\text{kp}}
= \frac{1}{N_{\kappa}^{\text{tot}}}
\sum_{(b,k)\in\mathcal{S}_{\text{batch}}}
\sum_{\kappa\in K^{\star}_{(b,k)}}
\Big[&
\lambda_{\text{loc}}\,
\mathcal{L}_{\text{reg}}\big(\hat{\boldsymbol{\kappa}}_{\text{loc}}(\kappa),\,\boldsymbol{\kappa}_{\text{loc}}\big)\notag\\
&+\,\lambda_{\text{type}}\,
\mathcal{L}_{\text{cls}}\big(\hat{\mathbf{p}}_{\text{type}}(\kappa),\,\kappa_{\text{type}}\big)
\Big],
\label{eq:kp_loss}
\end{align}
where \(N_{\kappa}^{\text{tot}}\) is the total number of ground-truth keypoints in the batch, and \(\lambda_{\text{loc}},\lambda_{\text{type}}\) are balancing weights.
The location regression term is the element-wise Smooth L1 loss
\begin{equation}
\mathcal{L}_{\text{reg}}(\hat{\mathbf{y}},\mathbf{y})
= \sum_{d}\mathrm{SmoothL1}(\hat{y}_{d}-y_{d}),
\label{eq:smoothl1_loss}
\end{equation}
with \(\mathrm{SmoothL1}(x)=0.5x^{2}\) if \(|x|<1\) and \(|x|-0.5\) otherwise. The type classification term is cross-entropy
\begin{equation}
\mathcal{L}_{\text{cls}}(\hat{\mathbf{p}},y_{\text{true}})
= -\log\!\big(\hat{p}_{y_{\text{true}}}\big),
\label{eq:cross_entropy_loss}
\end{equation}
where \(\hat{p}_{y_{\text{true}}}\) is the predicted probability of the ground-truth class.

\textbf{Trajectory alignment loss (\(\mathcal{L}_{\text{traj}}\)).}
To encourage consistency between the predicted macro-action sequence \(A'=\{a'_{k}\}_{k=1}^{N_{\text{seg}}}\) and the intended path encoded by the sketch, we compare a planar \(SE(2)\) rollout with a reference path sampled from \(S\).
From \(A'\) we generate a simulated pose sequence \(P_{\text{sim}}=\{p^{\text{sim}}_{t}\}_{t=0}^{N_{\text{sim}}}\) using a simple kinematic model: \texttt{forward} advances along the current heading by a fixed step; \texttt{turn\_p90}, \texttt{turn\_n90}, \texttt{turn\_p45}, \texttt{turn\_n45} update the orientation without translation; \texttt{check\_under} leaves the pose unchanged; \texttt{cover\_area} expands to a serpentine sweep within the annotated region (implemented by alternating straight runs and discrete turns as specified in Algorithm~\ref{alg:inference_loop_revised}).
The ground-truth path \(P_{\text{gt}}=\{p^{\text{gt}}_{u}\}_{u=0}^{N_{\text{gt}}}\) is obtained by resampling the sketched strokes in a common world frame.
We use Dynamic Time Warping (DTW) \cite{itakura2003minimum} over planar coordinates:
\begin{equation}
\label{eq:traj_loss}
\begin{split}
    \mathcal{L}_{\text{traj}} = \frac{1}{|\mathcal{S}_{\text{batch}}|} \sum_{b\in\mathcal{S}_{\text{batch}}} \mathrm{DTW} \Big( & \{(x^{\text{sim}}_{t},y^{\text{sim}}_{t})\}_{t=0}^{N^{(b)}_{\text{sim}}}, \\
    & \{(x^{\text{gt}}_{u},y^{\text{gt}}_{u})\}_{u=0}^{N^{(b)}_{\text{gt}}} \Big).
\end{split}
\end{equation}

\textbf{Total loss (\(\mathcal{L}_{\text{total}}\)).}
The overall objective combines task classification, keypoint supervision, and trajectory alignment:
\begin{equation}
\mathcal{L}_{\text{total}}
=\mathcal{L}_{\text{task}}
+\gamma\,\mathcal{L}_{\text{kp}}
+\beta\,\mathcal{L}_{\text{traj}},
\label{eq:total_loss_final}
\end{equation}
where \(\gamma\) and \(\beta\) balance the contributions of geometric supervision and path fidelity
(\(\gamma=0.1,\ \beta=0.05\)).
This multi-objective formulation promotes correct discrete decisions, geometric salience in the sketch encoder, and faithful physical execution of sketched paths.

\begin{algorithm*}[ht]
  \caption{Runtime inference and execution loop}
  \label{alg:inference_loop_revised}
  \begin{algorithmic}[1]
    \REQUIRE User input: image \(I\), sketch \(S=\{l_1,\dots,l_n\}\), optional language \(L\).
    \REQUIRE Robot platform \(\mathcal{M}\) with sensors providing perception \(P_t\).
    \REQUIRE Trained modules: \(\phi_S,\ \phi_V,\ \phi_L,\ \psi_{\text{fuse}},\ \pi_{\text{HL}}\).
    \REQUIRE Translator \(g_{\text{translate}}\); control params: \(L_{\text{max}}=0.5\) m,\ \(\theta_{\text{turn}}=30^{\circ}\),\ \(d_{\text{step}}=0.05\) m,\ \(d_{\text{safety}}=0.30\) m,\ \(h_{\text{clearance}}=1.00\) m.
    \STATE Preprocess \(S\): segment all strokes and concatenate into \(\mathcal{S}_{\text{seq}}=\{s_{1},\dots,s_{N_{\text{seg}}}\}\) using \(\theta_{\text{turn}}\) and \(L_{\text{max}}\).
    \FOR{\(k=1\) \TO \(N_{\text{seg}}\)}
      \STATE \(s_{\text{cur}} \leftarrow s_{k}\); update perception \(P_t\).
      \STATE \(F^{\text{fused}}_{k} \leftarrow \psi_{\text{fuse}}\big(\phi_S(s_{\text{cur}}),\ \phi_V(I),\ \phi_L(L)\big)\). \COMMENT{optionally conditioned on features derived from \(P_t\)}
      \STATE \(a'_{k} \leftarrow \arg\max_{a'\in\mathcal{A}_{\text{disc}}} P(a'\mid F^{\text{fused}}_{k};\theta_{\pi})\) \quad (Eq.~\ref{eq:command_prediction_final})
      \IF{\(a'_{k}=\texttt{forward}\)}
        \STATE \textit{obstacle} \(\leftarrow\) \textbf{false}; \ \textit{done} \(\leftarrow\) \textbf{false}; \ \textit{pose} \(\leftarrow\) GetRobotPose()
        \WHILE{not \textit{done} \textbf{ and } not \textit{obstacle}}
          \STATE \(a_k \leftarrow g_{\text{translate}}(\texttt{forward},\ \mathcal{M},\ d_{\text{step}})\)
          \STATE Execute \(a_k\); update \(P_t\), \textit{pose}
          \STATE \textit{obstacle} \(\leftarrow\) CheckForObstacleInPath\((P_t,\ d_{\text{safety}})\)
          \STATE \textit{done} \(\leftarrow\) HasReachedEndOfSegment\((s_{\text{cur}},\ \textit{pose})\)
        \ENDWHILE
        \IF{\textit{obstacle}}
          \STATE Execute \(g_{\text{translate}}(\texttt{halt},\ \mathcal{M})\)
          \STATE \textit{reachable\_under} \(\leftarrow\) CheckUnderObstacleRoutine\((P_t,\ \textit{obstacle},\ h_{\text{clearance}})\)
          \IF{\textit{reachable\_under}}
            \STATE Execute \(g_{\text{translate}}(\texttt{under\_obstacle\_maneuver},\ \mathcal{M})\)
          \ENDIF
          \STATE \textbf{continue} \COMMENT{proceed to next segment}
        \ENDIF
      \ELSIF{\(a'_{k}\in\{\texttt{turn\_p45},\texttt{turn\_n45},\texttt{turn\_p90},\texttt{turn\_n90}\}\)}
        \STATE Execute \(g_{\text{translate}}(a'_{k},\ \mathcal{M})\); wait for rotation completion
      \ELSIF{\(a'_{k}=\texttt{check\_under}\)}
        \STATE Focus sensors on the area indicated by \(s_{\text{cur}}\); update \(P_t\)
        \STATE Record CheckUnderObstacleRoutine\((P_t,\ \text{region}(s_{\text{cur}}),\ h_{\text{clearance}})\) \COMMENT{no primary motion}
      \ELSIF{\(a'_{k}=\texttt{cover\_area}\)}
        \STATE \(\mathcal{P}_{\text{serp}}\leftarrow\) GenerateSerpentinePlan\((s_{\text{cur}})\) \COMMENT{alternating straight runs and allowed discrete turns}
        \FOR{each macro-action \(u\in \mathcal{P}_{\text{serp}}\)}
          \IF{\(u=\texttt{forward}\)} 
            \STATE repeat lines 9–15 for the current sweep lane
          \ELSE
            \STATE Execute \(g_{\text{translate}}(u,\ \mathcal{M})\)
          \ENDIF
          \STATE update \(P_t\) and \textit{pose}
        \ENDFOR
      \ENDIF
    \ENDFOR
  \end{algorithmic}
\end{algorithm*}

\subsection{Training Procedure and Implementation Details}

Training is implemented in PyTorch \cite{paszke2019pytorch} on NVIDIA A100 GPUs (80\,GB). We use AdamW \cite{kingma2014adam} with weight decay \(1\times10^{-2}\) and initial learning rate \(5\times10^{-5}\). The schedule is cosine annealing with a linear warm-up over the first \(5\%\) of iterations, for a total of 100 epochs on the designated dataset.

Module initialization follows a fixed–trainable split. The visual encoder \(\phi_{V}\) is initialized from ImageNet-pretrained weights \cite{deng2009imagenet} and kept frozen. The language encoder \(\phi_{L}\) uses CLIP ViT-B/16 pretrained weights \cite{clip} and is also frozen. The sketch encoder \(\phi_{S}\), the fusion module \(\psi_{\text{fuse}}\), and the high-level policy \(\pi_{\text{HL}}\) are trainable and initialized with standard random schemes.

The training proceeds in two stages. First, the sketch encoder \(\phi_{S}\) is pre-trained in isolation using only sketch data with the keypoint supervision loss \(\mathcal{L}_{\text{kp}}\) (Eq.~\ref{eq:kp_loss}). This stage establishes a strong geometric prior in \(\phi_{S}\) before multimodal fusion. Second, the pretrained \(\phi_{S}\) is integrated with the fusion module \(\psi_{\text{fuse}}\) and the high-level policy \(\pi_{\text{HL}}\), and all trainable components are fine-tuned end-to-end by minimizing the composite objective \(\mathcal{L}_{\text{total}}\) (Eq.~\ref{eq:total_loss_final}), i.e., the weighted sum of task classification, keypoint supervision, and trajectory alignment with fixed \(\gamma=0.1\) and \(\beta=0.05\).

To improve robustness and generalization across users and scenes, we apply data augmentation during end-to-end training:
(1) Sketch augmentation: coordinate jitter (Gaussian, \(\sigma=1\) pixel per point) and small affine transforms (rotation \(\pm 5^{\circ}\), scaling \(\pm 10\%\)) on segments;
(2) Image augmentation: random resized crops (scale \([0.8,1.0]\), aspect ratio \([3/4,4/3]\)), random horizontal flips (probability \(0.5\)), and color jitter (brightness/contrast/saturation/hue up to \(0.2\));
(3) Language augmentation: not applied; we rely on the robustness of the frozen language encoder.

\subsection{Runtime Inference and Execution Procedure}
\label{ssec:inference}

During deployment, the system converts the multimodal instruction \(\mathcal{I}\) into an executable macro-action sequence through iterative interpretation and closed-loop control. The input is encoded and fused to form the runtime representation \(\mathcal{R}\), which conditions the high-level policy \(\pi_{\text{HL}}\). At execution time the robot queries perception \(P_t\) at each step to adapt to scene changes and obstacles. The overall procedure is summarized in Algorithm~\ref{alg:inference_loop_revised}.

Prior to inference, the sketch set \(S\) is deterministically segmented and linearized into a single ordered list of primitives
\(\mathcal{S}_{\text{seq}}\).
A new segment boundary is created when the local turning angle at a point triplet exceeds \(\theta_{\text{turn}}=30^{\circ}\) or when the current segment length exceeds \(L_{\text{max}}=0.5\,\text{m}\). When camera calibration is unavailable, a fixed pixel-length surrogate is used. Coordinates within each segment are normalized to \([-1,1]^2\). The sequence \(\mathcal{S}_{\text{seq}}\) provides the per-segment inputs processed by \(f_{\text{fuse}}\) and \(\pi_{\text{HL}}\) during runtime.

At runtime, the controller iterates over segments \(s_{k}\in\mathcal{S}_{\text{seq}}\).
For the current segment \(s_{k}\), the robot first acquires the latest perception \(P_t\) (e.g., RGB–D, LiDAR). Features derived from \(P_t\) may be incorporated to refresh the visual context.
The fusion module \(\psi_{\text{fuse}}\) then forms the fused representation
\(F^{\text{fused}}_{k}\) as in Eq.~\eqref{eq:fusion_revised} (optionally augmented with perception-derived features).
The high-level policy \(\pi_{\text{HL}}\) maps \(F^{\text{fused}}_{k}\) to a discrete macro-action \(a'_{k}\in\mathcal{A}_{\text{disc}}\) by Eq. \ref{eq:command_prediction_final}.
In the current implementation, \(\pi_{\text{HL}}\) depends on \(P_t\) only through its contribution to the fused features.

The predicted macro-action \(a'_{k}\) is translated into platform-specific low-level commands \(a_{k}\in\mathcal{A}_{\text{DoF}}\) by the translator \(g_{\text{translate}}\) for the target platform \(\mathcal{M}\):
\begin{itemize}
  \item Rotations (\texttt{turn\_p90}, \texttt{turn\_n90}, \texttt{turn\_p45}, \texttt{turn\_n45}): \(g_{\text{translate}}\) issues the corresponding angular displacement \(\Delta\theta\in\{+90^{\circ},-90^{\circ},+45^{\circ},-45^{\circ}\}\) to the motion controller.
  \item Forward progression (\texttt{forward}): motion is executed in increments of \(d_{\text{step}}=0.05\) m along the segment’s principal direction. After each increment the system updates perception \(P_t\) and tests for obstacles within a safety horizon \(d_{\text{safety}}=0.30\) m. The robot advances until the end of \(s_{k}\) is reached or an obstacle is detected.
  \item Obstacle handling during forward: upon detection, the robot halts and invokes \texttt{CheckUnderObstacleRoutine} on the obstacle region to estimate under-clearance. If the clearance exceeds \(h_{\text{clearance}}\), the space is treated as traversable and \(g_{\text{translate}}\) executes \texttt{under\_obstacle\_maneuver} (a predefined multi-DoF sequence: short advance, optional tool actuation, and retraction as needed). If not traversable, execution of \(s_{k}\) terminates and the controller proceeds to \(s_{k+1}\).
  \item Clearance query (\texttt{check\_under}): the same routine is executed on the area indicated by \(s_{k}\) to record reachability without issuing primary motion.
  \item Area coverage (\texttt{cover\_area}): \(g_{\text{translate}}\) expands the macro-action into a serpentine (zig–zag) plan composed of alternating \texttt{forward} runs and admissible discrete turns. As in the forward case, each straight run proceeds until reaching the local boundary or an obstacle; when an obstacle is handled, the sweep continues from the next lane.
\end{itemize}

The low-level commands \(a_{k}\in\mathcal{A}_{\text{DoF}}\) (e.g., velocity pairs \((v,\omega)\) for a mobile base, joint-space trajectories \(q(t)\) or end-effector twists for manipulators) are dispatched to the platform controllers via ROS \cite{ros}. For navigation, the ROS navigation stack consumes \((v,\omega)\) and performs local path following and obstacle avoidance. For manipulation, including under-obstacle maneuvers, MoveIt is used for motion planning and execution. The system monitors controller feedback and advances to the next segment only after successful completion or a guarded termination of the current command. This segment-by-segment execution, together with reactive obstacle handling and continuous perception checks, enables robust realization of sketched instructions in cluttered domestic environments.

\section{Experimental Setup}

\subsection{Experimental Environment and Hardware}

\begin{figure}[h]
    \centering
    \includegraphics[width=0.8\columnwidth]{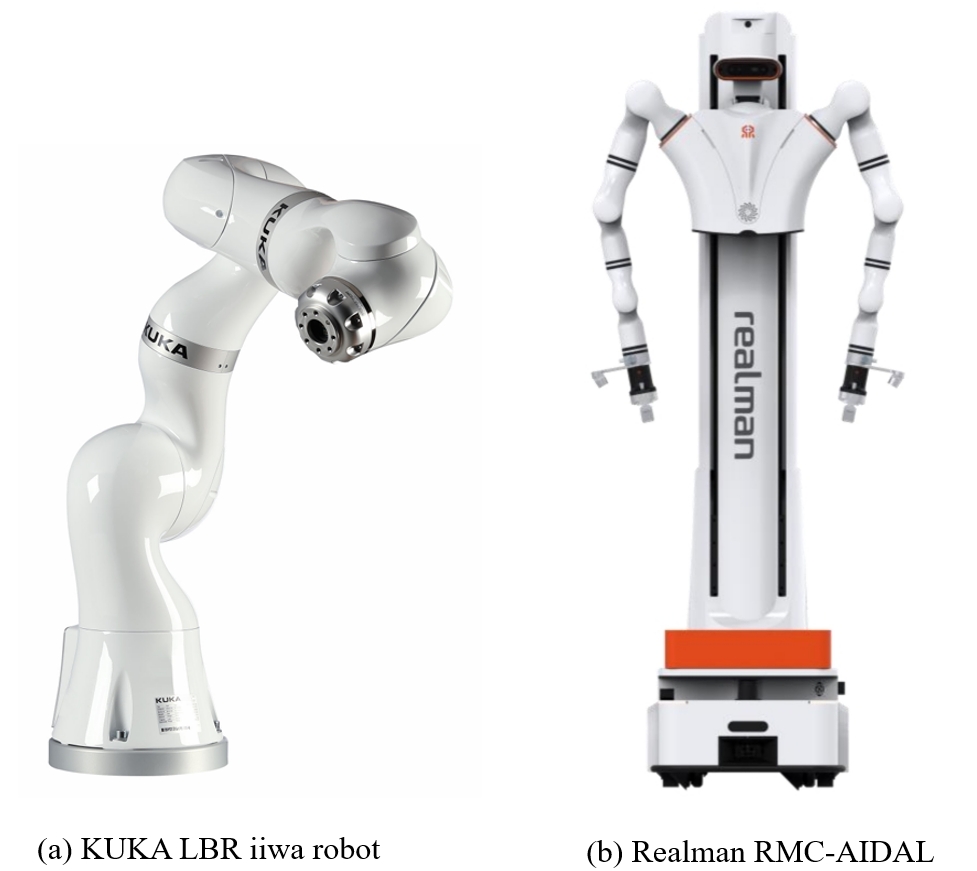}
    \caption{Representative robotic platforms relevant to this work. (a) The KUKA LBR iiwa, a 7-DoF collaborative manipulator. (b) The Realman RMC-AIDAL, a dual-arm mobile manipulation platform. }
    \label{fig:robot}
\end{figure}

To rigorously evaluate the performance and adaptability of AnyUser, we conducted studies in both simulated and real domestic settings, focusing on representative household tasks such as floor mopping and table wiping. For simulation, we used iGibson~2.0 \cite{li2021igibson} because it provides high-fidelity physics and photorealistic indoor scenes. Within this environment, the Freight mobile robot model served as the robotic agent, enabling controlled testing across diverse layouts derived from real-world scans and allowing systematic variation of scene complexity while holding robot embodiment constant.

For real-world validation, we selected hardware platforms that reflect the differing demands of the target tasks, as shown in Fig.~\ref{fig:robot}. Floor mopping, which requires integrated mobility and dual-arm coordination, was executed on the Realman RMC-AIDAL platform $(\mathcal{M}_{\text{realman}})$. The system comprises a differential-drive base with LiDAR navigation and obstacle avoidance, a central lifting column with 800~mm travel that extends the vertical workspace (approximately 200~mm to 2000~mm reach height, yielding an overall reach near 2.2~m), and two RM65-B 6-DoF arms. Each arm supports a 5~kg payload with $\pm 0.05$~mm repeatability (manufacturer specification). Perception is provided by three Intel RealSense D435 depth cameras and two RGB monitoring cameras, processed on an onboard NVIDIA Jetson AGX Orin (64~GB RAM, 1~TB SSD), and exposed to the control stack as real-time perception $P_t$. Each arm was equipped with an RMG24 parallel gripper (65~mm standard stroke, mass $\leq 0.5$~kg, rated load 4~kg) holding custom passive mopping tools. Experiments were conducted in laboratory spaces arranged to mimic typical apartments, with varied flooring and furniture configurations.

Table wiping, which demands precise manipulation in a confined workspace, was performed with a KUKA LBR iiwa 7 R800 CR collaborative arm. The arm provides 7~DoF, an 800~mm maximum reach, a 7~kg rated payload, and $\pm 0.1$~mm pose repeatability (ISO 9283), controlled via the KUKA Sunrise Cabinet. For these experiments the arm was statically mounted, though the platform supports floor, wall, or ceiling mounting. To supply visual context for task specification and to mimic a typical household camera viewpoint, an external Intel RealSense D435 was mounted overhead. The RGB stream from this camera served as the visual input $I$ for user interaction and scene understanding, consistent with settings where only visible-light imagery is available. The end-effector used a standard parallel gripper holding a custom wiping tool suitable for surface cleaning. Trials were conducted in controlled laboratory settings featuring varied table materials and representative clutter.

User interaction in both real-world setups employed a tablet interface that displayed a live or recently captured image $I$ from the relevant camera (either an onboard RMC-AIDAL camera or the external D435 for the KUKA setup). Participants, including researchers and naive users recruited to assess intuitiveness, specified tasks by sketching trajectories $S$ directly on $I$ and optionally adding brief language cues $L$. Execution and logging were managed with ROS \cite{ros}. All neural components of AnyUser ran on a dedicated workstation with an Intel Core i9 CPU, 64~GB RAM, and two NVIDIA RTX 4090 GPUs on Ubuntu 20.04. During real-time operation on the RMC-AIDAL, inference ran on the onboard Jetson AGX Orin.

\begin{table*}[h]
\centering
\caption{Quantitative performance evaluation on the HouseholdSketch dataset. Task length categories are defined by the number of detected corners (turning points) in the input sketch: \textbf{Short} ($\leq$ 2 corners), \textbf{Medium} (3-5 corners), and \textbf{Long} ($\geq$ 6 corners). The table presents Single-Step Success Rate \textbf{(SSSR)}, Single-Step Strict Path Adherence Rate \textbf{(SSSPAR)}, Full Task Completion Rate \textbf{(FTCR)}, and Full Task Strict Path Adherence Rate \textbf{(FTSPAR)}, reported in percentage (\%). Results are disaggregated by scene type and task length category.}
\label{tab:success-rates}
\resizebox{1\textwidth}{!}{
\fontsize{8pt}{10pt}\selectfont
\begin{tabular}{|c|c|c|c|c|c|c|c|c|c|c|c|c|}
\hline
\multicolumn{1}{|c|}{} & \multicolumn{3}{c|}{Single-Step} & \multicolumn{3}{c|}{Single-Step Strict Path} & \multicolumn{3}{c|}{Full Task} & \multicolumn{3}{c|}{Full Task Strict Path} \\
\multicolumn{1}{|c|}{Scene} & \multicolumn{3}{c|}{Success Rate (\%)} & \multicolumn{3}{c|}{Adherence Rate (\%)} & \multicolumn{3}{c|}{Completion Rate (\%)} & \multicolumn{3}{c|}{Adherence Rate (\%)} \\
\cline{2-13}
 & Short & Medium & Long & Short & Medium & Long & Short & Medium & Long & Short & Medium & Long \\ \hline
Bedroom      & 85.3 & 86.1 & 85.7 & 78.1 & 78.6 & 78.3 & 78.0 & 64.5 & 50.8 & 66.2 & 51.1 & 36.4 \\ \hline
Kitchen      & 83.8 & 84.5 & 84.1 & 76.9 & 77.2 & 77.0 & 74.3 & 60.2 & 46.5 & 62.7 & 48.5 & 33.8 \\ \hline
Living room  & 87.4 & 87.0 & 87.2 & 79.5 & 79.2 & 79.4 & 79.1 & 65.7 & 52.3 & 67.5 & 52.3 & 37.5 \\ \hline
Bathroom     & 84.6 & 85.0 & 84.8 & 77.4 & 77.7 & 77.5 & 76.5 & 62.3 & 48.9 & 64.8 & 49.2 & 35.1 \\ \hline
Corridor     & 83.2 & 83.5 & 83.1 & 75.8 & 76.1 & 75.9 & 72.8 & 58.1 & 44.6 & 60.4 & 46.1 & 31.6 \\ \hline
Staircase    & 82.7 & 83.0 & 82.5 & 74.9 & 75.2 & 74.8 & 70.4 & 55.6 & 42.1 & 57.9 & 43.7 & 29.8 \\ \hline
Region       & 83.5 & 83.9 & 83.6 & 75.6 & 75.9 & 75.7 & 71.6 & 56.3 & 43.0 & 59.1 & 44.9 & 30.9 \\ \hline
Other room   & 84.0 & 84.4 & 84.1 & 76.7 & 77.0 & 76.8 & 73.9 & 59.0 & 45.7 & 61.5 & 47.2 & 32.7 \\ \hline
All & 84.3 & 84.7 & 84.4 & 76.9 & 77.1 & 76.9 & 74.6 & 60.2 & 46.7 & 62.5 & 47.9 & 33.5 \\ \hline
\end{tabular}
}
\end{table*}

\subsection{Experimental Procedure}
Our evaluation combined simulated trials in iGibson~2.0 \cite{li2021igibson} with the Fetch Freight mobile base \cite{wise2016fetch} and real-world deployments that covered floor mopping with the Realman RMC-AIDAL and table wiping with the KUKA LBR iiwa. The procedures tested whether the system can correctly interpret user instructions and execute tasks robustly across varied settings.

In simulation, each trial began by loading a selected household environment and rendering a static viewpoint image $I$ that represents the user perspective. Simulated user input $(S,L)$ was either programmatically generated or drawn through an interface, specifying paths for navigation or regions for surface coverage. The AnyUser inference pipeline in Algorithm~\ref{alg:inference_loop_revised} was executed end to end. Ground truth state from the simulator was used to detect collisions, verify task completion by checking whether the robot trajectory or tool trace covered the target region implied by the sketch, and log performance metrics.

For real-world experiments, each trial began with configuring the task environment (e.g., positioning the RMC-AIDAL at its start pose or arranging tabletop objects for the KUKA arm). An operator or participant captured or selected the scene photograph $I$ via the tablet interface. The user then sketched trajectories $S$ directly on the image and, when desired, provided a concise language cue $L$. After confirmation, AnyUser automatically preprocessed $S$, decomposing it into an ordered sequence of geometric primitives
$\mathcal{S}_{\text{seq}}=\{s_{1},\dots,s_{N_{\text{seg}}}\}$
using a curvature threshold of $>30^{\circ}$ and a maximum projected segment length of approximately $0.5\,\text{m}$.

During execution the system may incorporate real-time perception \(P_t\) as an image-channel input alongside the original \((I,S,L)\). Concretely, when enabled, we form \((I,S,L,P_t)\) and encode \(P_t\) with the same frozen visual backbone \(\phi_V\) used for \(I\). All visual frames, whether the initial third-person photograph \(I\) or the egocentric robot view \(P_t\), are resized to \(224\times224\) and normalized with ImageNet statistics to match \(\phi_V\). For reproducibility, we emphasize that in this release \(P_t\) is RGB from the robot camera; depth and LiDAR are used by the platform’s safety and navigation controllers but are not fed to \(\psi_{\text{fuse}}\). In our internal experiments (reported in the Appendix) where depth was injected into \(\psi_{\text{fuse}}\), the depth map was linearly scaled to \([0,255]\) and replicated to three channels before ViT preprocessing; this did not outperform RGB without additional modality-specific training, so the final model defaults to RGB only. Because \(I\) and \(P_t\) arise from different viewpoints, we do not attempt explicit geometric warping between them. Instead, sketches are grounded in \(I\) for intent, and \(P_t\) provides local, egocentric evidence for reactive checks. In fusion, \(P_t\) contributes through a simple late-fusion extension of Eq.~\eqref{eq:fusion_revised}:
\begin{equation}
F^{\text{fused}}_{k}=\mathrm{MLP}_{\text{fuse}}\!\big([\,f^{S}_{k};\, f^{V}_{\text{att}}(I);\, f^{V}_{\text{cls}}(I);\,
f^{L};\, \eta_t\, f^{V,\text{live}}_{\text{cls}}(P_t)\,]\big),
\end{equation}
where \(\eta_t\in\{0,1\}\) gates the contribution of live perception. In our implementation \(\eta_t=1\) only for obstacle handling, including \texttt{check\_under} and post-detection clearance assessment, and \(\eta_t=0\) otherwise, which leaves \((I,S,L)\) as the dominant drivers of macro-action selection. In all configurations, the language encoder always receives a non-empty text input: when users provide a description, we concatenate it with a fixed system prompt $L_0$, and when users are non-verbal or user language is ablated, we supply only $L_0$ (reported in the Appendix).

The system then entered the iterative execution loop in Algorithm~\ref{alg:inference_loop_revised}, processing segments sequentially. For each segment $s_k$, the multimodal interpreter fused $\phi_S(s_k)$, $\phi_V(I)$, and $\phi_L(L)$, optionally augmented by real-time perception $P_t$ from onboard sensors, to produce $F^{\text{fused}}_{k}$. The high-level policy $\pi_{\text{HL}}$ predicted the most probable macro-action $a'_{k} \in \mathcal{A}_{\text{disc}}$ (Eq.~\ref{eq:command_prediction_final}), which was passed to the platform-specific translation module $g_{\text{translate}}$ for execution.

Execution behavior depended on the predicted macro-action. For rotational macros \(\{\texttt{turn\_p45},\ \texttt{turn\_n45},\ \texttt{turn\_p90},\ \texttt{turn\_n90}\}\), \(g_{\text{translate}}\) generated target angular displacements \(\Delta\theta\in\{\pm45^{\circ},\,\pm90^{\circ}\}\) that were executed by ROS controllers. For \texttt{forward}, \(g_{\text{translate}}\) initiated stepwise motion with step length \(d_{\text{step}}=0.05\,\text{m}\) along the segment direction. After each step the system updated perception \(P_t\) and performed obstacle checking using depth sensors within a safety field of \(d_{\text{safety}}=0.30\,\text{m}\) ahead. Motion continued until the estimated end of \(s_k\) was reached or an obstacle was detected. Upon obstacle detection the robot halted and invoked \texttt{CheckUnderObstacleRoutine}, which analyzed \(P_t\) in the obstacle region to estimate under-clearance. If the clearance exceeded the platform threshold \(h_{\text{clearance}}=1.00\,\text{m}\), the space was considered traversable and \(g_{\text{translate}}\) generated an \texttt{UnderObstacleManeuver} comprising a short sequence of multi-DoF commands to proceed under the obstacle; otherwise the segment was marked locally obstructed and execution advanced to \(s_{k+1}\). When the predicted macro was \texttt{check\_under}, the same perception routine was executed on the area indicated by \(s_k\) without locomotion. The resulting low-level commands \(a_k\in\mathcal{A}_{\text{DoF}}\) (e.g. \((v,\omega)\) for the RMC-AIDAL base or joint velocities \(\dot{q}\) for the arms) were dispatched via ROS topics to the appropriate controllers. The ROS navigation stack handled base motion and MoveIt handled arm planning and execution. The system waited for controller feedback signaling completion before proceeding to \(s_{k+1}\).

An entire task specified by the initial sketch \(S\) was considered successful if the robot executed the sequence associated with all segments \(\mathcal{S}_{\text{seq}}\) without unrecoverable errors or manual intervention beyond the initial instruction. A human operator recorded success or failure for each trial while observing execution. For the large-scale household deployments referenced in the Introduction, trials comprised complex multi-segment tasks across different rooms or surface conditions, with the environment sometimes undergoing minor changes between instruction and execution. Safety was enforced through software-defined workspace limits and velocity caps within the ROS control loops. We also reported the parameterization and practical configuration in the Appendix.

\subsection{Evaluation Metric} \label{metric}

To provide a comprehensive assessment of AnyUser, we employ a suite of human–evaluated success metrics. Because user intent is expressed through free-form sketches, automated scoring alone can miss important nuances. Trained evaluators therefore assess outcomes under a standardized protocol that captures both goal achievement and execution fidelity. We report four primary metrics:

\textbf{Full Task Completion Rate (FTCR).} FTCR measures overall usefulness from the end-user perspective. It is the ratio of tasks successfully completed to the total number of tasks attempted. A task is marked successful if the robot achieves the primary objective implied by \(\mathcal{I}=(I,S,L)\) (e.g., mopping the designated floor area or wiping the specified table region) without operator intervention due to system failure, unrecoverable errors, or critical deviation from the intended goal. Minor departures from the sketched path do not affect FTCR.

\textbf{Full Task Strict Path Adherence Rate (FTSPAR).} FTSPAR measures end-to-end geometric fidelity. It is the ratio of tasks that are both completed (per FTCR) and executed with high fidelity to the geometry encoded by the user sketches \(S\), relative to all attempted tasks. Evaluators apply predefined tolerances to the executed trajectory with respect to the segment sequence \(\mathcal{S}_{\text{seq}}\) derived from \(S\) (e.g., a lateral band of \(\pm 25\,\text{cm}\) for floor mopping and \(\pm 5\,\text{cm}\) for table wiping around the intended path).

\textbf{Single-Step Success Rate (SSSR).} SSSR probes robustness at the segment level. It is the ratio of sketch segments \(s_k\in\mathcal{S}_{\text{seq}}\) whose corresponding macro-action \(a'_k\in\mathcal{A}_{\text{disc}}\) is executed successfully, to the total number of processed segments. Success means that the predicted action (e.g., \texttt{forward}, \texttt{turn\_p90}, \texttt{check\_under}) completes its subtask without immediate failure or safety violation.

\textbf{Single-Step Strict Path Adherence Rate (SSSPAR).} SSSPAR quantifies geometric accuracy for individual segments. It is the ratio of successfully executed segments that also adhere strictly to their intended geometry, to the total number of successfully executed segments. Adherence is judged using local tolerances, such as small lateral deviation for \texttt{forward} motions and small angular error for \texttt{turn\_*} actions, relative to the geometry of \(s_k\).

Together, FTCR captures goal attainment, FTSPAR assesses global path fidelity, SSSR diagnoses per-segment robustness, and SSSPAR evaluates local geometric precision. This multi-layered, human-standardized evaluation provides a balanced view of capability and reliability in realistic domestic scenarios.

\begin{figure*}[h]
    \centering
    \includegraphics[width=\linewidth]{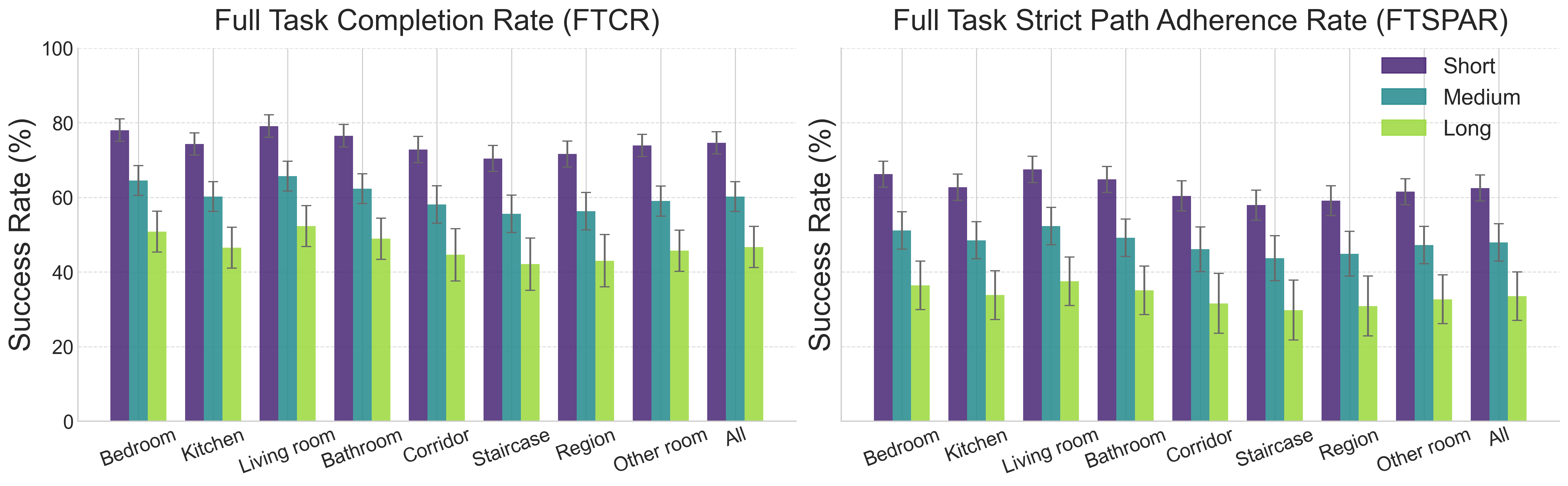}
    \caption{Scene-specific task-level performance comparison. Task length categories are defined by sketch complexity (Short: $\leq$ 2 corners, Medium: 3-5 corners, Long: $\geq$ 6 corners). The figure presents (Left) Full Task Completion Rate (FTCR) and (Right) Full Task Strict Path Adherence Rate (FTSPAR) in percentage (\%). Results are shown for each scene category from the HouseholdSketch dataset, grouped by task length. Error bars depict simulated standard error, indicating expected performance variability.}
    \label{fig:scene_performance}
\end{figure*}

\begin{figure}[h]
    \centering
    \includegraphics[width=\columnwidth]{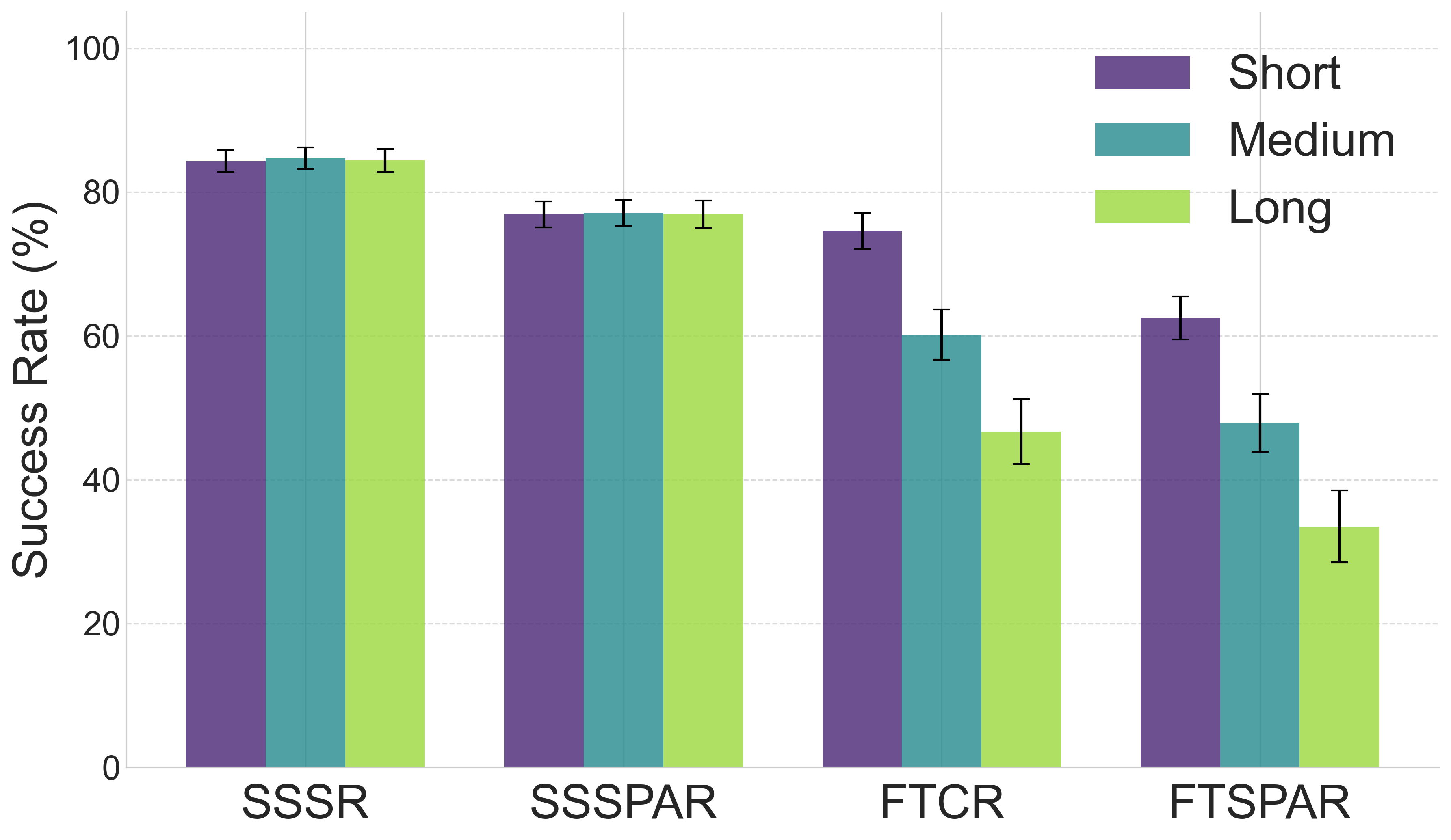}
    \caption{Aggregate performance comparison across metrics and task lengths. Task length categories (Short: $\leq$ 2 corners, Medium: 3-5 corners, Long: $\geq$ 6 corners) are based on the number of turns in the input sketch. Bars represent the average success rates (\%) for SSSR, SSSPAR, FTCR, and FTSPAR, computed across all scenes from the HouseholdSketch dataset. Error bars indicate simulated standard error, reflecting anticipated experimental variability.}
    \label{fig:overall_performance}
\end{figure}

\begin{figure*}[h]
    \centering
    \includegraphics[width=\linewidth]{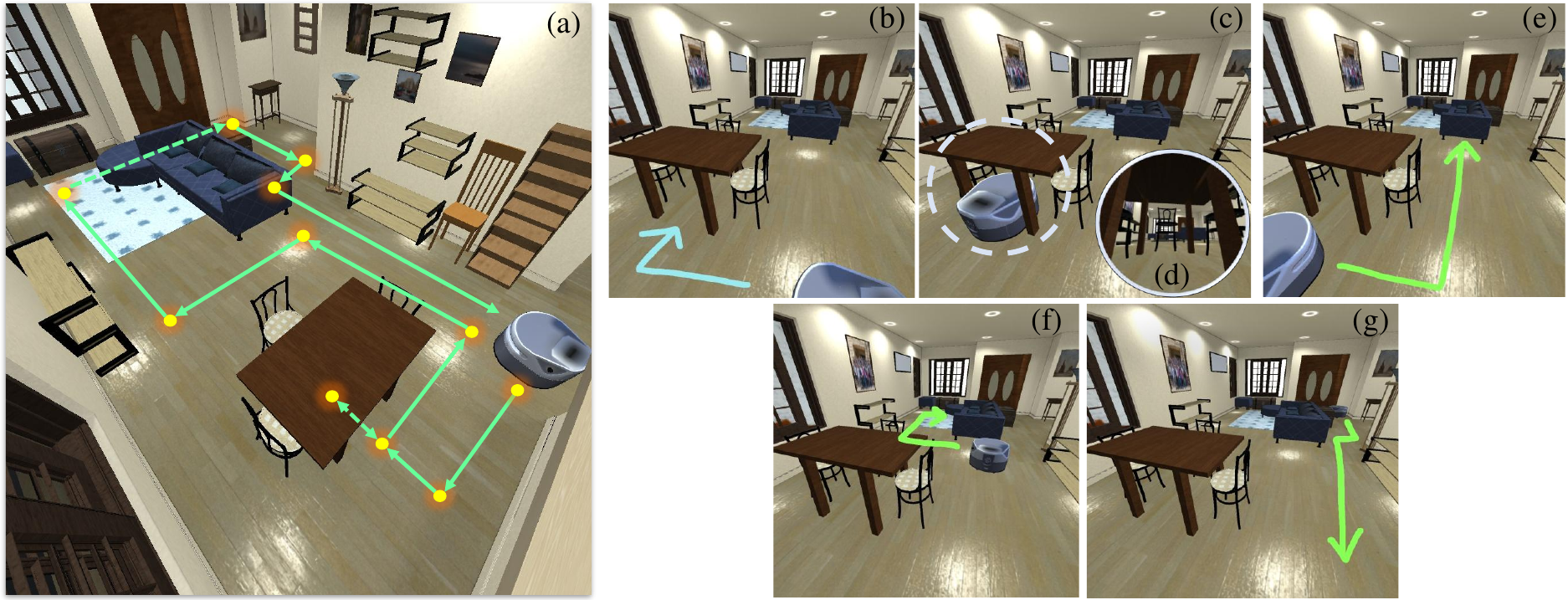}
    \caption{Qualitative illustration of system operation in the iGibson simulation environment. (a) User-provided sketch overlaid on the scene, defining a multi-segment navigation task around furniture and under a table (keypoints highlighted). (b)-(g) Sequence depicting the robot's execution trace: (b) initial forward motion, (c) approaching the table triggering under-obstacle check, (d) simulated perception check under the table, (e)-(f) navigating turns and proceeding under the table, (g) final segment execution. The sequence demonstrates task decomposition, path following, and integration of adaptive behaviors like obstacle checking.}
    \label{fig:case}
\end{figure*}

\section{Experimental Results}
\subsection{Evaluation on HouseholdSketch}

The system was evaluated quantitatively on the HouseholdSketch dataset, with results summarized in Table~\ref{tab:success-rates},~\ref{tab:ablation} and visualized through aggregate trends in Fig.~\ref{fig:overall_performance} and scene-specific outcomes in Fig.~\ref{fig:scene_performance}. All main reported results use the standard zero-shot prompt configuration for the language channel. For analysis, tasks were grouped by sketch complexity based on the number of detected corners in the input sketch: \textbf{Short} tasks contain \(\leq 2\) corners, \textbf{Medium} tasks contain \(3\!-\!5\) corners, and \textbf{Long} tasks contain \(\geq 6\) corners. A qualitative illustration of system behavior in interactive simulation is provided in Fig.~\ref{fig:case}.

\subsubsection{Main Result}

Analysis of the Single-Step Success Rate (SSSR), which quantifies successful execution of the intended macro-action \(a'_k\in\mathcal{A}_{\text{disc}}\) for each sketch segment \(s_k\in\mathcal{S}_{\text{seq}}\), indicates strong foundational reliability. As shown in the leftmost group of bars in Fig.~\ref{fig:overall_performance}, the average SSSR is consistently high at approximately \(84.4\%\) and remains stable across Short, Medium, and Long tasks. The compact error bars further suggest low variance, implying that the mechanism for interpreting elementary geometric inputs and triggering the corresponding discrete commands (e.g., \texttt{forward}, \texttt{turn\_p90}) operates dependably regardless of overall task length. This consistent per-segment performance forms a solid basis for the system’s interaction capabilities.

However, assessing execution precision reveals the practical difficulty of translating geometric intent into exact robot motion, even in simulation that mirrors real-world challenges. The Single-Step Strict Path Adherence Rate (SSSPAR), shown in Fig.~\ref{fig:overall_performance} (second group), averages lower at approximately \(76.9\%\) yet remains stable across task lengths. The persistent gap between SSSR and SSSPAR indicates that the system often selects the correct macro-action type while strict geometric conformity to the sketched segment is less frequent. In operational terms, a command such as \texttt{forward} may produce minor lateral drift relative to an ideal straight line, and a \texttt{turn\_p90} command may result in a rotation slightly different from exactly \(90^{\circ}\). These effects are consistent with control inaccuracies or state-estimation imperfections that also arise on physical platforms.

The cumulative impact of these per-segment characteristics is apparent in end-to-end outcomes. Fig.~\ref{fig:overall_performance} shows a clear decline in both Full Task Completion Rate (FTCR) and Full Task Strict Path Adherence Rate (FTSPAR) as task length increases. The average FTCR drops from \(74.6\%\) for Short tasks to \(46.7\%\) for Long tasks, indicating a reduced probability of achieving the overall objective as complexity grows. FTSPAR exhibits an even larger decrease, from \(62.5\%\) for Short tasks to \(33.5\%\) for Long tasks. The visibly larger error bars for Medium and Long tasks further suggest increasing variance in outcomes for more complex instructions, consistent with the compounding of small execution errors and deviations over longer sequences derived from \(\mathcal{S}_{\text{seq}}\).

Disaggregating results by environmental context, as shown in Fig.~\ref{fig:scene_performance}, provides additional insight. The trend of performance degradation with increasing task length holds across diverse scene types, from Bedroom to Staircase. The figure also quantifies the influence of the environment on execution success. Navigationally simpler or less cluttered settings such as Living room tend to yield higher FTCR and FTSPAR than more constrained scenes such as Corridor or Staircase, which pose greater challenges for path planning and execution. The side-by-side view of FTCR (left) and FTSPAR (right) highlights the performance gap introduced by strict path adherence criteria, indicating that functional goal achievement is more attainable than achieving the same goal with precise adherence to the sketched path.

\begin{table*}[t]
\centering
\caption{Ablation on input modalities and live perception on HouseholdSketch (overall across scenes and task lengths). Numbers are percentages. In parentheses are absolute changes relative to the full model.}
\label{tab:ablation}
\begin{tabular}{lcccc}
\hline
Setting & SSSR & SSSPAR & FTCR & FTSPAR \\
\hline
$I{+}S{+}L{+}P_t$ \ (full)       & 84.6 & 77.0 & 60.9 & 48.4 \\
$I{+}S{+}L$ \ (no $P_t$)         & 84.0 \ ({-}0.6) & 76.0 \ ({-}1.0) & 58.7 \ ({-}2.2) & 47.2 \ ({-}1.2) \\
$I{+}S$ \ (no user $L$)          & 83.8 \ ({-}0.8) & 76.4 \ ({-}0.6) & 59.4 \ ({-}1.5) & 47.1 \ ({-}1.3) \\
$S{+}L$ \ (no $I$)               & 78.1 \ ({-}6.5) & 70.2 \ ({-}6.8) & 41.3 \ ({-}19.6) & 28.4 \ ({-}20.0) \\
$I{+}L$ \ (no $S$)               & 68.5 \ ({-}16.1) & 55.3 \ ({-}21.7) & 32.1 \ ({-}28.8) & 20.2 \ ({-}28.2) \\
\hline
\end{tabular}
\end{table*}

\subsubsection{Case Study}
Qualitative examples from the interactive iGibson environment (Fig.~\ref{fig:case}(a)) shows a representative multi-segment sketch (green polyline with yellow keypoints) instructing the robot to navigate around a couch and under a dining table within a living room. (b)-(g) depict the resulting execution: the robot follows \(\mathcal{S}_{\text{seq}}\) segment by segment, performing \texttt{forward} motions (b, e, g) and turn commands (e, f) consistent with the sketch geometry. (c), (d) illustrate handling of an implicit interaction derived from the path. As the sketched route leads under the table (c, dashed circle), the system invokes the \texttt{check\_under} routine. (d) provides a simulated first-person view used to assess traversability beneath the table. After confirming clearance in this example, the robot continues along the intended segments (f, g). This case study demonstrates decomposition of intricate sketches into executable macro-actions and integration of perceptual checks and adaptive behaviors during runtime.

\subsubsection{Ablation study}
To quantify the contribution of each input channel and the optional live perception, we conducted ablations on HouseholdSketch. We report results aggregated over all scenes and task lengths. The full model uses the photograph, sketch, and language tuple \((I,S,L)\) with the default system prompt (reported in the Appendix) in the language channel and incorporates live perception \(P_t\) into the fusion pathway. When we ablate the user language, the language encoder still receives the fixed system prompt that defines the macro-action vocabulary and safety priors. When we ablate the image, the visual input is replaced by a heavily blurred constant image, which preserves tensor shape but removes semantic content. When we ablate the sketch, only \((I,L)\) are provided to the network. Table~\ref{tab:ablation} summarizes Single-Step Success Rate (SSSR), Single-Step Strict Path Adherence Rate (SSSPAR), Full Task Completion Rate (FTCR), and Full Task Strict Path Adherence Rate (FTSPAR).

The sketch channel is the primary driver of spatial intent. Removing \(S\) causes the largest degradation across all metrics, with SSSR dropping by 16.1\% and FTCR by 28.8\%. This confirms that free-form strokes provide the strongest geometric prior for \(\pi_{\text{HL}}\). Removing \(I\) produces the next largest drop, reflecting the importance of visual–semantic grounding for associating strokes with scene surfaces, detecting keep-out regions, and aligning segment directions to traversable space. By contrast, ablating user language while retaining the fixed system prompt yields small reductions in SSSR and SSSPAR (below one point) and modest decreases in FTCR and FTSPAR. This is consistent with language serving chiefly to refine intent and resolve ambiguity rather than to supply the core spatial scaffold.

Incorporating live perception \(P_t\) yields modest gains on the aggregate, with improvements of +0.6\% SSSR and +2.2\% FTCR. As expected given our design, the benefit concentrates on segments that encounter obstacles and on longer tasks that traverse occluded areas. Stratified analysis shows that for Long tasks the gains rise to \(\Delta\)FTCR \(= {+}3.2\)\% and \(\Delta\)FTSPAR \(= {+}2.1\)\%, while for Short tasks the improvements are within 1\%. This aligns with the role of \(P_t\) in the current implementation, which primarily informs under-obstacle checks and local halting behavior when unexpected obstacles are detected.

\subsection{Evaluation on Real Robot}

To validate the practical applicability and robustness of AnyUser beyond simulation, we conducted experiments on two distinct real-world robotic platforms detailed in Section V-A: the KUKA LBR iiwa 7-DoF collaborative arm for manipulation tasks, and the Realman RMC-AIDAL dual-arm mobile manipulator for tasks requiring navigation and manipulation. These experiments focused on executing representative domestic tasks specified via user sketches in laboratory environments mimicking home settings.

\begin{figure}[t] 
    \centering
    \includegraphics[width=\columnwidth]{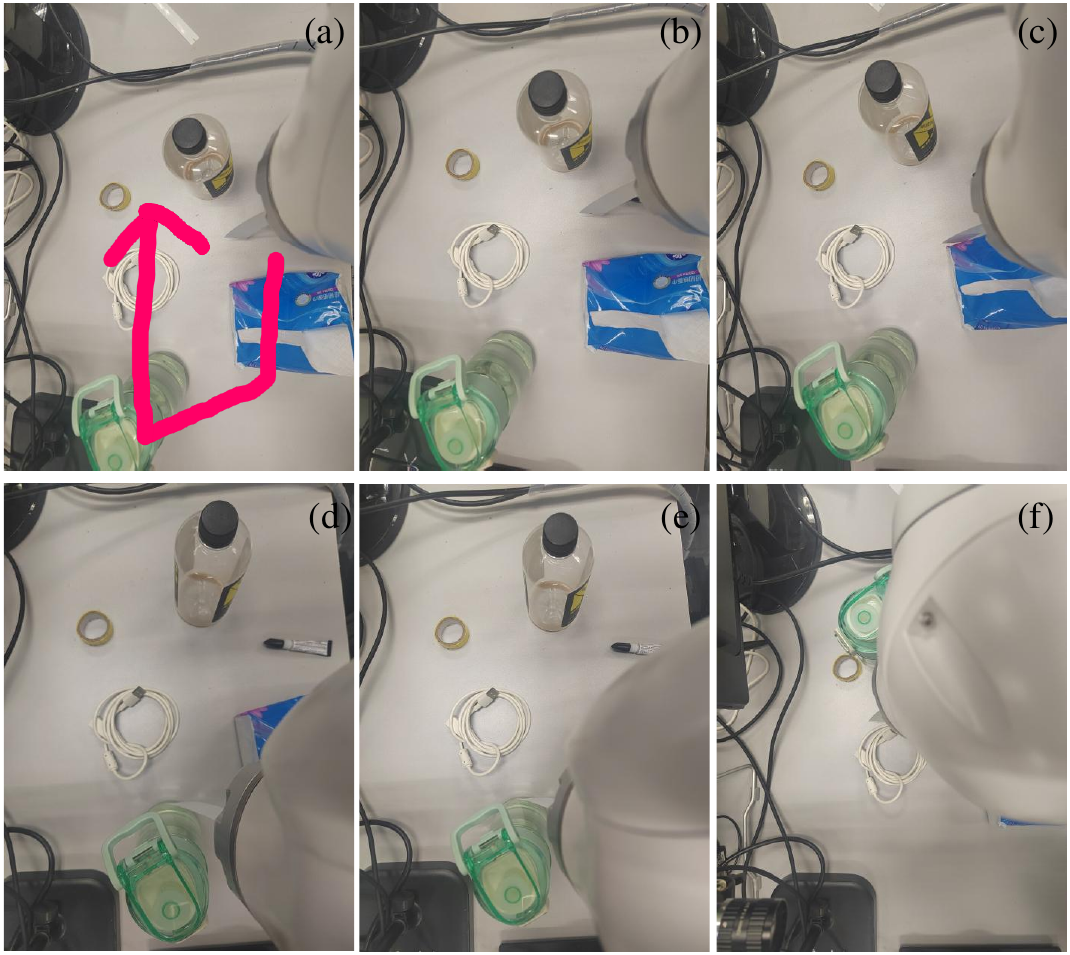} 
    \caption{Qualitative illustration of system operation with the KUKA LBR iiwa robot executing a table wiping task. (a) User-provided sketch overlaid on the camera view, indicating a path (arrow) and a target area (rectangle). (b)-(f) Sequence depicting the robot's execution: (b) Initial state. (c)-(d) Arm approaches the target region. (e) Arm executes wiping motion along the sketched path. (f) Arm interacts with the designated rectangular area. The sequence demonstrates successful 3D grounding, trajectory execution, and task completion based on the sketch.}
    \label{fig:kuka_exp} 
\end{figure}

\subsubsection{KUKA LBR iiwa (7-DoF Arm)} We evaluated the system's ability to interpret sketches for precise manipulation tasks, such as wiping specific areas on a cluttered tabletop. Fig.~\ref{fig:kuka_exp} illustrates a typical trial. The user provides an initial image of the scene with overlaid sketches (Fig.~\ref{fig:kuka_exp}(a)), in this case, indicating both a path (arrow) to approach a target region near the center and a rectangular area near the tissue packet to be wiped. The system successfully grounds this 2D instruction in the 3D workspace, accounting for the camera perspective. Fig.~\ref{fig:kuka_exp}(b)-(f) show snapshots of the execution sequence: the robot plans and executes a collision-free trajectory to approach the target (Fig.~\ref{fig:kuka_exp}(c)-(d)), performs the wiping motion along the specified path trajectory (Fig.~\ref{fig:kuka_exp}(e)), and addresses the designated rectangular area (Fig.~\ref{fig:kuka_exp}(f) shows the arm operating in that vicinity). This successful execution demonstrates several key capabilities in a real-world setting: (1) accurate 3D grounding of free-form 2D sketches from a static image; (2) interpretation of compound sketches involving different geometric primitives (path and area) corresponding to different actions; (3) generation and execution of precise, collision-aware manipulation trajectories in a moderately cluttered environment; and (4) successful task completion based on the user's sketched intent. Performance in these trials qualitatively supports the high success rates observed in simulation and user studies for manipulation-oriented tasks.

\begin{figure}[t] 
    \centering
    \includegraphics[width=\columnwidth]{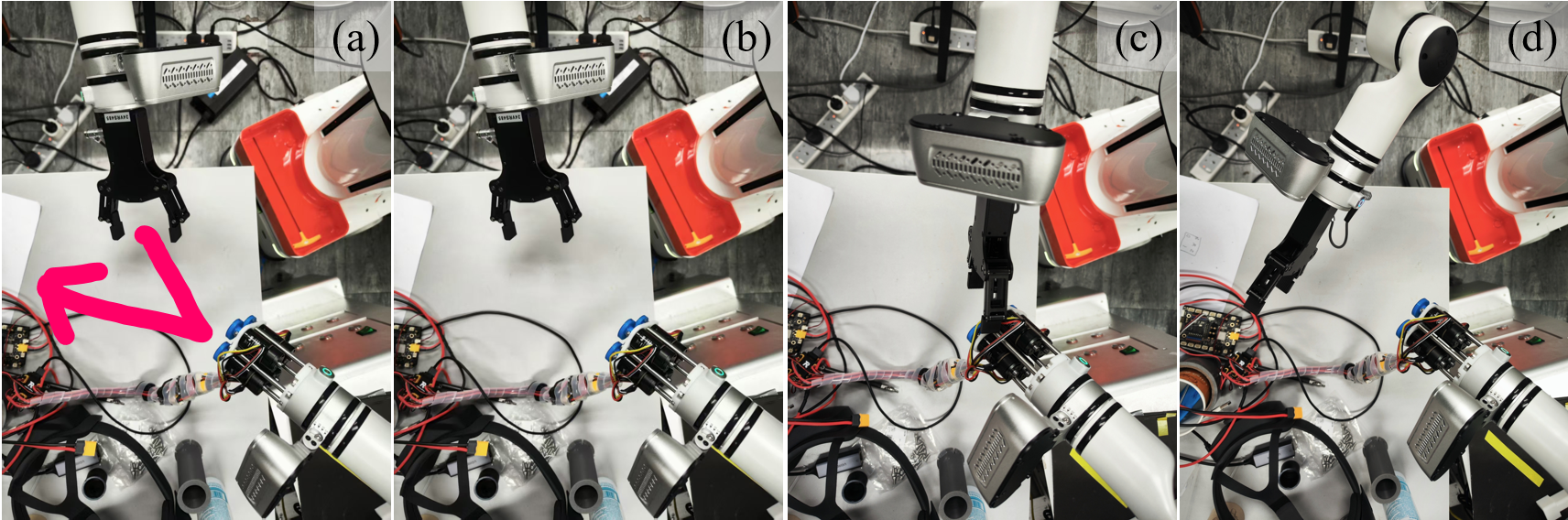}
    \caption{Qualitative illustration of dual-arm cover-area task on the Realman RMC-AIDAL mobile manipulator driven by sketch-based commands. (a) User sketch overlaid on the onboard camera image: a single curved arrow on the tabletop specifies the region to be traversed and the desired sweeping direction. After sketch-guided navigation brings the base to a pose in front of the table,  (b)–(d) the hierarchical policy instantiates a \texttt{cover\_area} macro that converts the arrow into synchronized, collision-aware trajectories for both RM65-B arms: the right arm performs the primary sweeping motion, while the left arm adopts a reactive tucked posture to maintain clearance in the cluttered workspace.
    }
    \label{fig:realman_exp}
\end{figure}

\subsubsection{Realman RMC-AIDAL (Dual-Arm Mobile Manipulator)} 
On the Realman RMC-AIDAL platform, we evaluate AnyUser on dual-arm mobile manipulation tasks in which both the base pose and the two arms are commanded from sketches. For the trial shown in Fig.~\ref{fig:realman_exp}, the user first uses the same sketch-based navigation interface as in the previous experiments to guide the mobile base to a pose in front of a cluttered tabletop. The user then draws a single sweeping arrow over the tabletop region (Fig.~\ref{fig:realman_exp}(a)), which specifies a \texttt{cover\_area} behavior: the surface to be traversed and the desired sweeping direction. The fused sketch–image representation is grounded into a 3D description of this surface, and the hierarchical policy instantiates a \texttt{cover\_area} macro that generates synchronized, collision-aware trajectories for both RM65-B arms. 

In this cluttered setup, the system employs an asymmetric coordination strategy to prevent mutual interference. The right arm acts as the primary executor, tracing the specified sweeping region. Meanwhile, commanding the left arm to move ``away'' could risk collisions with surrounding hardware; therefore, the left arm executes a reactive ``tucking'' motion to maintain a compact anchor posture. Because Fig.~\ref{fig:realman_exp} displays the onboard egocentric camera view, the 2D projection creates an optical illusion that the left arm is moving dangerously close to the right arm. In reality, the dual-arm planner allows the left arm's joints to yield space for the right arm, navigating precisely along the boundary of the dynamic collision hull while strictly maintaining a predefined 3D safety clearance.

Snapshots of the execution are shown in Fig.~\ref{fig:realman_exp}(b)–(d): starting from the sketched entry point, the two arms coordinate to move the tool along the arrow direction while maintaining clearance from each other and from surrounding cables and hardware. For clarity, the navigation sketch and motion are not visualized in this figure; the panels focus on the dual-arm cover-area stage. These experiments indicate that the same sketch vocabulary used for single-arm manipulation and pure navigation also suffices to drive integrated mobility and coordinated dual-arm behaviors on a high-DoF platform, without changing the user interface or requiring task-specific programming.

\begin{table*}[t]
\centering
\caption{User Study Performance Metrics Across Diverse Demographic Groups. Values represent mean (± standard deviation) where applicable. TCSR is presented as a percentage. Subjective ratings and EFR are on a 1-5 scale (higher is better).}
\label{tab:user_study_results}
\begin{tabular}{lcccc}
\hline
Metric & Elderly (n=10) & Sim. Non-Verbal (n=8) & Low Tech Literacy (n=7) & Control (n=7) \\
\hline
Specification Time (s) & 62.5 (±14.8) & 55.1 (±10.2) & 68.3 (±16.5) & 49.8 (±9.0) \\
Attempts per Task & 1.5 (±0.7) & 1.3 (±0.5) & 1.7 (±0.8) & 1.2 (±0.4) \\
Full Task Completion Rate (FTCR) (\%) & 90.0\% & 93.8\% & 85.7\% & 96.4\% \\
Expert Fidelity Rating (EFR) (1-5) & 4.0 (±0.8) & 4.2 (±0.6) & 3.8 (±0.9) & 4.4 (±0.5) \\
\hline
Subjective Ease of Use (1-5) & 4.3 (±0.7) & 4.5 (±0.5) & 3.9 (±1.0) & 4.6 (±0.4) \\
Subjective Confidence (1-5) & 4.1 (±0.8) & 4.4 (±0.6) & 3.7 (±1.1) & 4.5 (±0.5) \\ %
Subjective Satisfaction (1-5) & 4.2 (±0.7) & 4.6 (±0.4) & 4.0 (±0.9) & 4.7 (±0.3) \\
\hline
\end{tabular}
\end{table*}

\subsection{User Study with Diverse Demographics}

To validate the core objective of AnyUser in democratizing robot interaction, particularly for populations often underserved by complex technological interfaces, we conducted a focused user study involving diverse demographic groups. Recognizing the potential of assistive robotics to benefit elderly individuals and those with communication impairments, our study specifically recruited participants representing these target populations, alongside individuals with varying levels of technical literacy and no prior programming experience. The primary aim was to assess the system's intuitiveness, efficiency, and effectiveness in enabling these users to specify and oversee common domestic tasks using the AnyUser interface.

The study involved \textit{N=32} participants, carefully selected and categorized into four groups: elderly adults (n=10, aged 65-80 years, with varying degrees of experience with modern technology), individuals simulating non-verbal communication (n=8, instructed to rely solely on sketches without verbal clarification during task specification to simulate conditions like aphasia or mutism), users self-reporting low technical literacy (n=7, identified via a pre-study questionnaire assessing comfort and frequency of use with digital devices), and a control group with general technical familiarity but no specific robotics or programming background (n=7). All participants provided informed consent under an institutional review board (IRB) approved protocol. Experiments were conducted in a laboratory environment configured to resemble a domestic kitchen and living area, utilizing the Realman RMC-AIDAL for floor cleaning tasks (``Mop the area around the table'') and the KUKA LBR iiwa for table wiping tasks (``Wipe this section of the table, avoiding the cup''). Participants interacted with the AnyUser system via a standard tablet displaying a recently captured image from the relevant robot camera. Each session began with a standardized tutorial explaining the sketching interface and the optional use of minimal language cues (except for the non-verbal group). Participants were then asked to specify and initiate the execution of the two predefined tasks. We collected several metrics: (1) Task Specification Time, measured from the moment the task was presented to the user confirming their instruction; (2) Number of Attempts, recording how many times a user modified or restarted their sketch before confirming; (3) Full Task Completion Rate (FTCR), assessed by an expert observer according to the definition in Sec.~\ref{metric}, determining if the robot successfully achieved the primary goal of the task (e.g., target area cleaned/wiped) without safety violations or critical failures requiring manual intervention; (4) User Subjective Feedback, gathered post-study using a 5-point Likert scale questionnaire focusing on perceived ease of use, confidence in the instruction provided, perceived system understanding of their intent, and overall satisfaction; (5) Expert Fidelity Rating (EFR), where the expert observer rated the congruence between the robot's executed path/actions and the geometric intent conveyed by the user's sketch on a 1-5 scale (1=poor, 5=excellent alignment).

The empirical results derived from the real-world user study, presented in Table~\ref{tab:user_study_results}, lend strong quantitative and qualitative support to the accessibility and effectiveness of the AnyUser system across a spectrum of user capabilities. Crucially, the Full Task Completion Rate (FTCR) remained high for all participant categories, achieving 90.0\% for elderly users, 93.8\% for those simulating non-verbal interaction, and 85.7\% for users with low technical literacy, compared to 96.4\% for the control group. This demonstrates the system's core capability to translate user intent into successful task execution, even for individuals who might typically struggle with complex interfaces. While the FTCR for the low technical literacy group was the lowest, achieving success in over 85\% of tasks still signifies substantial usability. Efficiency metrics reveal expected variations: the average Task Specification Time was longest for the low technical literacy group (68.3s ± 16.5s) and the elderly group (62.5s ± 14.8s), compared to the control group (49.8s ± 9.0s). Similarly, these groups required slightly more Attempts per Task on average (1.7 ± 0.8 and 1.5 ± 0.7, respectively, versus 1.2 ± 0.4 for control). However, the key finding is that these moderate increases in specification time and attempts did not fundamentally impede the ability of these users to ultimately convey their intent successfully, as evidenced by the high FTCR scores. This suggests the sketching paradigm is sufficiently intuitive to allow users to self-correct and converge on a functional instruction without excessive frustration or failure. The performance of the simulated non-verbal group is particularly noteworthy; they achieved the second-highest FTCR (93.8\%) with specification times (55.1s ± 10.2s) closer to the control group, strongly validating the sketch modality's power as a primary channel for conveying precise spatial intent when verbal communication is restricted.

Subjective feedback reinforces the positive objective findings. Perceived Ease of Use was rated highly across groups, averaging 4.3 (±0.7) for elderly, 4.5 (±0.5) for non-verbal, and 4.6 (±0.4) for control participants. Even the low technical literacy group reported a positive average score of 3.9 (±1.0), indicating general usability despite potentially less familiarity with tablet interfaces. Confidence that the system understood the instruction followed a similar pattern, with non-verbal users reporting high confidence (4.4 ± 0.6), potentially due to the directness of the sketch modality. The slightly lower average confidence score (3.7 ± 1.1) and higher standard deviation for the low technical literacy group might suggest a greater variability in user experience or residual uncertainty about the technology within this cohort, aligning with their slightly lower FTCR and higher attempt rate. Nonetheless, overall satisfaction remained high, averaging above 4.0 for all groups, with the control and non-verbal groups reporting the highest satisfaction (4.7 ± 0.3 and 4.6 ± 0.4, respectively). Furthermore, the Expert Fidelity Rating (EFR), assessing the geometric faithfulness of the robot's execution to the sketch, averaged 4.0 or higher for the elderly, non-verbal, and control groups, indicating good alignment. The slightly lower EFR for the low technical literacy group (3.8 ± 0.9) suggests that while tasks were often completed successfully (85.7\% FTCR), the precision of execution relative to the sketch might have been slightly reduced, potentially correlating with less precise initial sketches from this group. Collectively, these findings, situated within the context of real-world robotic task execution, provide compelling evidence that AnyUser effectively bridges the usability gap. The system successfully lowers the barrier for robot instruction, making sophisticated robotic capabilities accessible and controllable for users regardless of their technical expertise, age, or communication abilities, thereby substantiating its potential for broad societal impact in assistive and domestic applications. 

\textbf{Non-verbal users vs. Ablations.} The simulated non-verbal condition in the user study corresponds to participants who provide only sketches \(S\) and omit user language. Technically, this does not remove the language channel from the architecture. The text encoder still receives a fixed system prompt (reported in the Appendix) that enumerates the macro-action vocabulary and safety priors, exactly as in the \(I{+}S\) ablation in Table~\ref{tab:ablation}. No architectural changes are required to support non-verbal operation, and \(\psi_{\text{fuse}}\) handles the absent user text by encoding only the system prompt. The high FTCR observed for non-verbal participants in real robot trials reflects differences in domain and protocol relative to HouseholdSketch: participants specified a smaller set of tasks on specific platforms with curated scenes and received brief training on the sketch interface, while the ablation aggregates across diverse simulated homes and a wide range of task lengths. The two evaluations therefore quantify complementary aspects of the system: the ablation isolates channel contributions at scale, and the user study measures end-to-end usability when users rely on sketching alone.

\section{Discussion}

\subsection{Long-horizon performance: error taxonomy and compounding effects.}
Our quantitative analysis reveals a gap between the high accuracy of segment-wise action selection (SSSR) and the reduced success under strict geometric adherence (SSSPAR, FTSPAR), with the aggregate impact most visible on long tasks where FTCR drops as sequence length grows. To better understand failure modes on Long tasks in HouseholdSketch, we manually audited a stratified sample of failures and categorized primary causes. Table~\ref{tab:long_error_breakdown} summarizes the breakdown.

\begin{table}[t]
\centering
\caption{Primary failure sources on Long tasks in HouseholdSketch (audited subset, percentages sum to 100).}
\label{tab:long_error_breakdown}
\begin{tabular}{l r}
\hline
Failure source & Share \% \\
\hline
Perception-driven mismatches in grounding with \(I\) or \(P_t\) & 29.1 \\
Sketch-to-scene misalignment in fusion \(f_{\text{fuse}}\) & 23.7 \\
Macro-action misclassification in \(\pi_{\text{HL}}\) & 20.8 \\
Execution drift in \(g_{\text{translate}}\) and low-level control & 19.6 \\
Timeouts or conservative halts from safety checks & 6.8 \\
\hline
\end{tabular}
\end{table}

\begin{table*}[h]
\centering
\caption{Effect of action-space resolution on Long tasks in HouseholdSketch. Percentages reported; deltas are absolute changes relative to the baseline.}
\label{tab:action_resolution}
\begin{tabular}{lcccc}
\hline
Setting & SSSR & SSSPAR & FTCR & FTSPAR \\
\hline
Baseline: \(\pm45^{\circ}, \pm90^{\circ}\), \(0.05\,\mathrm{m}\) & 84.4 & 76.9 & 46.7 & 33.5 \\
Coarser: \(\pm90^{\circ}\), \(0.10\,\mathrm{m}\) & 83.9 & 71.8 \,(\(-5.1\)) & 42.3 \,(\(-4.4\)) & 28.1 \,(\(-5.4\)) \\
Finer: \(\pm22.5^{\circ}, \pm45^{\circ}, \pm90^{\circ}\), \(0.025\,\mathrm{m}\) & 84.1 & 79.3 \,(+2.4) & 45.6 \,(\(-1.1\)) & 35.4 \,(+1.9) \\
Continuous yaw regression, \(0.05\,\mathrm{m}\) & 84.0 & 78.4 \,(+1.5) & 44.1 \,(\(-2.6\)) & 34.2 \,(+0.7) \\
\hline
\end{tabular}
\end{table*}

Failures are not uniformly distributed across a trajectory. A survival-style analysis over segment index shows a clear compounding pattern: 17.3\% of failures occur in the first third of segments, 24.6\% in the middle third, and 58.1\% in the final third. This is consistent with small pose errors and local mis-groundings accumulating over time, particularly when the route traverses occluded regions or passes under furniture that triggers additional checks based on \(P_t\).

At the representation level, we observed two recurrent phenomena. First, keypoint drift on elongated strokes can slightly bias the estimated heading of \(s_k\), which in turn affects the selection between \texttt{turn\_p45} versus \texttt{turn\_p90} before a forward advance. Second, ambiguous local context can cause segment features to attend to visually similar patches in \(F^{V}_{\text{grid}}\), especially in scenes with repetitive texture. At the control level, rare but impactful wheel slip in simulation and controller quantization can produce small yaw errors that are not fully corrected unless a new turn macro is explicitly emitted by \(\pi_{\text{HL}}\).

\subsection{Does action-space resolution matter?}
We investigated whether changing the macro-action resolution affects long-horizon outcomes. Table~\ref{tab:action_resolution} reports an ablation on Long tasks comparing the baseline action set (turns of \(\pm 45^{\circ}\) and \(\pm 90^{\circ}\); step length \(d_{\text{step}}{=}0.05\,\mathrm{m}\)) with coarser, finer, and partially continuous variants. Results indicate the expected trade-off between expressivity and horizon length. Finer angular resolution improves FTSPAR but creates longer sequences that are more exposed to compounding errors, slightly reducing FTCR. Coarser resolution shortens sequences but degrades adherence. Regressing a continuous yaw angle increases per-step adherence on simple corners but introduces higher variance, reducing FTCR when scenes are cluttered.

These findings support the current design choice. A compact discrete vocabulary yields stable training, predictable safety envelopes, easier platform transfer through \(g_{\text{translate}}\), and interpretable logs for post hoc analysis. Smoothness is recovered at execution time by the platform controllers, which already operate in continuous velocity or joint spaces. A promising direction is a hybrid scheme in which \(\pi_{\text{HL}}\) emits discrete macros along with small continuous residuals that are bounded by safety constraints.

\subsection{Mitigations for long-horizon reliability.}
Several principled improvements follow from the above analysis. First, periodic re-anchoring of segments to live cues can reduce drift. This can be implemented as light-weight ICP of the current egocentric image onto local patches of the authoring photograph \(I\), followed by a small heading correction before advancing on \(\texttt{forward}\) macros. Second, uncertainty-aware fusion can down-weight ambiguous visual regions in \(\psi_{\text{fuse}}\) by exposing a confidence head that modulates the logits of \(\pi_{\text{HL}}\). Third, keypoint re-detection at segment boundaries can prevent the propagation of early curvature errors. Fourth, look-ahead obstacle checks using \(P_t\) two steps ahead of the current \(s_k\) can preemptively schedule corrective turns instead of hard halts. Finally, closed-loop residual control in \(g_{\text{translate}}\) can apply a bounded lateral PID correction during \texttt{forward} to keep the base centered on the intended stroke vector.

\subsection{Broader applicability of sketch-based input.}
Although our experiments focus on floor mopping and table wiping, sketching is a general medium for communicating spatial intent. On planar surfaces, many everyday tasks admit the same representation: cleaning shower walls, wiping windows, erasing whiteboards, and polishing refrigerator doors can all be specified by drawing one or more arrows or loops on a single photograph of the surface. In each case, the sketch defines a coverage region and preferred sweeping direction in image space, which AnyUser grounds to the corresponding surface frame and then executes using the same \texttt{cover\_area} macro and serpentine planning logic used for floors and tables.

Beyond strictly planar motion, sketches can guide richer behaviors while keeping the user interface unchanged. Arrows drawn along object handles can encode grasp approach vectors that the system translates into approach and grasp sequences. Freehand 2D masks drawn in one or more views can be fused into 3D exclusion regions that define keep-out volumes for the planner. Tool paths over moderately curved surfaces can be indicated by layering strokes from two viewpoints, which are lifted to a simple surface model before execution. For articulated tasks such as opening a cabinet, a short arc around the handle can indicate the desired rotation about the hinge, which AnyUser can map to a combined approach and pull macro. Extending AnyUser to these settings primarily requires augmenting the translator \(g_{\text{translate}}\) with a small number of task-specific controllers and macros while reusing the same fused representation \(\mathcal{R}\) and multimodal understanding pipeline.

\subsection{Interpretability and the discrete vs. continuous trade-off.}
Discretizing the high-level action space brings three concrete benefits. It stabilizes learning by simplifying the target space for \(\pi_{\text{HL}}\). It improves safety and platform transfer since every macro maps to a finite, auditable set of low-level behaviors through \(g_{\text{translate}}\). It preserves human interpretability: logs of \(a'_k\) can be reviewed alongside the sketch to diagnose failure. The trade-off is reduced expressivity at the decision layer, which we mitigate by relying on continuous low-level control to smooth motion and by choosing a sufficiently fine step length \(d_{\text{step}}\). The resolution ablation in Table~\ref{tab:action_resolution} indicates that further refining the turn set yields small adherence gains but does not singularly solve long-horizon degradation, which is dominated by compounding perception and alignment errors. A hybrid policy that emits discrete macros with constrained continuous residuals is therefore a natural next step.

\subsection{Operational model and future extensions.}
Our current operational model relies on a single authoring photograph \(I\) with live perception \(P_t\) used for local checks and halts. This design keeps interaction lightweight and map-free. It is less equipped for large environmental changes outside the current field of view. Integrating optional persistent mapping, global change detection, and lightweight replanning would address this limitation while preserving the sketch-centric interface. The task domain can be broadened by coupling \(\mathcal{R}\) with grasp planners and physics-aware manipulation controllers, particularly for dexterous or bimanual tasks. Finally, error detection and recovery can be generalized beyond obstacle checks by equipping the system with anomaly detectors that monitor force, vision, and progress metrics, triggering recovery macros within the same hierarchical framework.

\section{Conclusion}

This paper introduced AnyUser, a unified framework enabling intuitive robot instruction through free-form sketches on environmental images, augmented optionally by language, without requiring prior maps or models. By integrating multimodal instruction understanding with a hierarchical control policy, AnyUser translates non-expert user intent into spatially grounded, executable robot actions, significantly lowering the barrier for specifying complex tasks in unstructured domestic settings. Extensive evaluations demonstrated its effectiveness: quantitative benchmarks on the large-scale HouseholdSketch dataset confirmed high accuracy in interpreting sketch semantics (average Single-Step Success Rate $\approx$ 84.4\%); real-world experiments on both manipulator (KUKA LBR iiwa) and mobile manipulator (Realman RMC-AIDAL) platforms validated practical applicability for tasks like wiping and area coverage; and comprehensive user studies (N=32) across diverse demographics confirmed high task completion rates (85.7\%-96.4\%) and usability, proving its accessibility. AnyUser provides a robust and highly accessible paradigm for human-robot interaction, establishing a strong foundation for future research aimed at enhancing the precision, scope, and resilience of assistive robots operating collaboratively with humans in everyday environments.

\bibliographystyle{IEEEtran}
\bibliography{ref}

\section{Parameterization and Practical Configuration}
\label{supp:params}

This section documents the rationale and practical configuration of the parameters used in Algorithm~1 (in the main paper). The values were selected to balance robustness, safety, and responsiveness across heterogeneous homes and platforms, while remaining simple to reproduce. All symbols follow the notation in Table~1 (in the main paper).

\subsection{Sketch length threshold \(L_{\text{max}}=0.5\,\text{m}\)}
This threshold limits the maximum straight segment produced by the sketch segmentation. It serves two purposes: it regularizes hand-drawn strokes into motion-sized primitives for \(\pi_{\text{HL}}\), and it bounds the distance traveled per primitive so that reactive checks can update frequently. The value \(0.5\,\text{m}\) was chosen to match typical domestic maneuver lengths for a mobile base operating near furniture edges and for wiping strokes on tabletops. It yielded stable macro-action prediction and reduced over-segmentation due to small drawing jitters.

\textbf{Pixel-to-meter mapping.} When camera intrinsics and an approximate planar support are available, we estimate a homography \(\mathbf{H}_{g\!\to\!i}\) from the ground (or table) plane to the image using four coplanar keypoints and compute the local metric scale for a 2D image displacement \(\Delta \mathbf{u}\) by
\[
\Delta \mathbf{x}_{\text{world}} \approx \big(\mathbf{J}_{\mathbf{H}^{-1}}(\mathbf{u}) \, \Delta \mathbf{u}\big)_{xy},
\]
where \(\mathbf{J}_{\mathbf{H}^{-1}}(\cdot)\) is the Jacobian of the inverse homography and \((\cdot)_{xy}\) projects to the plane coordinates. Segment splitting then uses the world length \(\|\Delta \mathbf{x}_{\text{world}}\|\) relative to \(L_{\text{max}}\).

When camera parameters are unknown, we use a calibrated pixel proxy that is resolution and field-of-view agnostic:
\[
L_{\text{max}}^{\text{px}} \;=\; \kappa \,\sqrt{H^2+W^2},
\]
with \(\kappa=0.08\) for \(224\times224\) inputs, which corresponds to \(\sim 25\) pixels per maximum segment. This \(\kappa\) was obtained by matching the median image footprint of \(0.5\,\text{m}\) in our collection of third-person photographs that cover a room-scale field of view. In practice, implementers may set \(\kappa\in[0.06,0.10]\) and optionally apply a post-pass that merges consecutive segments if the odometry-measured travel between split points is below \(0.20\,\text{m}\). This keeps the behavior stable when the photograph is unusually close or wide.

\subsection{Turn threshold \(\theta_{\text{turn}}=30^{\circ}\)}
Curvature-based splitting inserts a boundary when the instantaneous turning angle exceeds \(\theta_{\text{turn}}\). The threshold is aligned with the discrete action set in Sec.~IV-B4 (in the main paper), which includes \(45^{\circ}\) and \(90^{\circ}\) rotations. A \(30^{\circ}\) cut reliably captures intentional corners while filtering sketch noise. We compute turning angles over a short window of ordered points and apply angle hysteresis of \(5^{\circ}\) to avoid chatter near the threshold.

\subsection{Execution step length \(d_{\text{step}}=0.05\,\text{m}\)}
Forward motion is issued in fixed steps. A \(5\,\text{cm}\) stride provides responsive closed-loop updates at typical controller rates of \(10\!-\!30\,\text{Hz}\) and keeps the robot within the obstacle safety envelope between checks. Steps larger than \(0.10\,\text{m}\) reduced obstacle reaction quality; steps smaller than \(0.03\,\text{m}\) increased controller overhead without measurable gains in adherence.

\subsection{Obstacle safety distance \(d_{\text{safety}}=0.30\,\text{m}\)}
The look-ahead distance for obstacle checks is set from a conservative stopping-distance calculation,
\[
d_{\text{stop}} \approx \frac{v^2}{2a_{\text{brake}}} + v\,t_{\text{latency}} + \delta_{\text{sensor}},
\]
with \(v=0.30\,\text{m/s}\), \(a_{\text{brake}}=0.60\,\text{m/s}^2\), \(t_{\text{latency}}=0.10\,\text{s}\), and \(\delta_{\text{sensor}}=0.10\,\text{m}\) for depth noise and missed returns around specular or dark surfaces. This yields \(d_{\text{stop}}\approx 0.25\,\text{m}\). We use \(0.30\,\text{m}\) to add margin across platforms and floors.

\subsection{Under-obstacle clearance \(h_{\text{clearance}}=1.00\,\text{m}\)}
This threshold decides whether the \texttt{check\_under} routine will authorize an \texttt{UnderObstacleManeuver}. It is platform-dependent and should exceed the effective vertical envelope of the end-effector and tool during the sweep:
\[
h_{\text{clearance}} \;\ge\; h_{\text{tool}}(\alpha) + \epsilon,
\]
where \(h_{\text{tool}}(\alpha)\) is the maximum height the tool occupies at the sweep pitch \(\alpha\), and \(\epsilon\) accounts for depth error and deflection. For the Realman RMC-AIDAL with the passive mopping attachment and the nominal sweep pitch used in our trials, \(h_{\text{tool}}\approx 0.85\,\text{m}\); using \(\epsilon\approx 0.15\,\text{m}\) gives the default \(1.00\,\text{m}\). For tabletop wiping with the KUKA LBR iiwa, where the arm does not attempt under-obstacle insertion, this check is rarely triggered and \(h_{\text{clearance}}\) can be reduced accordingly.

\IEEEpubidadjcol

\subsection{Translator \(g_{\text{translate}}\)}
The translator maps macro-actions to low-level commands using platform kinematics and native controllers. It does not affect sketch segmentation. Tuning \(L_{\text{max}}\), \(\theta_{\text{turn}}\), and \(d_{\text{step}}\) changes only the frequency and granularity of macro-actions supplied to \(g_{\text{translate}}\).

\subsection{Recommended tuning recipe}
If intrinsics are known and a dominant plane is visible, prefer metric splitting via homography. If not, start with \(\kappa=0.08\) for \(L_{\text{max}}^{\text{px}}\), keep \(\theta_{\text{turn}}=30^{\circ}\), set \(d_{\text{step}} \in [0.04,0.06]\,\text{m}\) to match controller rate and braking performance, and set \(d_{\text{safety}}\ge d_{\text{stop}}+0.05\,\text{m}\). Choose \(h_{\text{clearance}}\) from the tool envelope using the formula above. These defaults reproduce the behaviors reported in our experiments while remaining adaptable to different cameras and robots.

\section{Default Language Prompt for Segment-Level Macro-Action Selection}
\label{app:prompt}

For non-verbal operation and for standardizing the language channel during inference, we use a fixed textual prior \(L_0\) that encodes the action set, safety constraints, and decision rules, as shown in Listing~\ref{lst:anyuser_prompt}. At runtime, the policy consumes one sketch segment \(s_k\) at a time and produces the macro-action for that segment. The same \(L_0\) is also used in ablations where user-provided language is omitted, ensuring that \((I,S,L_0)\) remains informative.

\begin{figure*}[t]
  \centering
  \begin{minipage}{0.95\textwidth}
\begin{lstlisting}[style=anyuserprompt, caption={Default segment-level prompt $L_0$ used by AnyUser. Curly-brace fields are filled by the runtime preprocessor for the current segment \(s_k\).}, label={lst:anyuser_prompt}]
[ROLE]
You are a segment-to-macro-action planner for a household robot.
Decide the macro-actions that best realize the current sketch segment s_k
in a cluttered, real home, subject to safety and platform constraints.

[ACTION VOCABULARY]
Choose only from:
{ forward, turn_p45, turn_n45, turn_p90, turn_n90, check_under, cover_area }.

[INPUTS FOR THIS SEGMENT]
Segment index: {SEGMENT_INDEX} of {N_SEG}
Flags: is_path={IS_PATH}, is_area={IS_AREA}, is_closed={IS_CLOSED}
Geometric stats (meters unless noted):
  length={LENGTH_M}, signed heading change dpsi_deg={DELTA_YAW_DEG},
  mean_curvature={MEAN_CURV}, corner_count={N_CORNERS}
Scene priors from photograph I:
  under_table_prior={UNDER_TABLE_PRIOR in [0,1]},
  traversable_prior={TRAVERSABLE_PRIOR in [0,1]}
Live perception gate eta_t (0 or 1): {ETA_T}
If eta_t=1 (obstacle handling context):
  obs_ahead (within d_safety=0.30m)={OBS_AHEAD}
  estimated under-clearance h_est_m={H_CLEAR_M or "unknown"}
Global control params:
  L_max=0.50m, theta_turn=30deg, d_step=0.05m, h_clearance=1.00m

[DECISION RULES]
1) Area segments:
   If is_area=true or is_closed=true -> output ["cover_area"] only.

2) Turns for path segments (by |dpsi_deg|):
   If >= 67.5deg -> 90deg turn: turn_p90 if dpsi_deg>0 else turn_n90.
   If in [22.5deg, 67.5deg) -> 45deg turn: turn_p45 if dpsi_deg>0 else turn_n45.
   Else -> prefer "forward" (see Rule 3).

3) Forward for path segments:
   Prefer ["forward"] when |dpsi_deg|<22.5deg and is_path=true.
   When eta_t=0, ignore obstacle/clearance fields (use I-based priors).

4) Obstacle-under check (only if eta_t=1):
   If obs_ahead=true and h_est_m="unknown" -> ["check_under"].
   If obs_ahead=true and h_est_m < h_clearance -> do NOT output "forward";
      instead choose the best turn per Rule 2.
   If obs_ahead=true and h_est_m >= h_clearance -> "forward" allowed.

5) One macro per path segment (controller handles d_step stepping).
   Do not invent parameters.

6) Safety:
   Use only the allowed tokens. For conflicting priors, choose the safest
   action that still progresses the segment intent.

[OUTPUT FORMAT]
Return exactly:
{
  "macro_actions": ["<one of: forward | turn_p45 | turn_n45 | turn_p90 | turn_n90 | check_under | cover_area>"],
  "confidence": <float in [0,1]>
}

[EXAMPLES]
A) Straight path, no live perception:
   Inputs: is_path=true, is_area=false, dpsi_deg=3, eta_t=0
   Output: {"macro_actions":["forward"], "confidence":0.92}

B) Right-angle corner:
   Inputs: is_path=true, is_area=false, dpsi_deg=-88, eta_t=0
   Output: {"macro_actions":["turn_n90"], "confidence":0.95}

C) Approach under table with live perception, unknown clearance:
   Inputs: is_path=true, is_area=false, dpsi_deg=2, eta_t=1,
           obs_ahead=true, h_est_m="unknown"
   Output: {"macro_actions":["check_under"], "confidence":0.88}

D) Area polygon:
   Inputs: is_area=true, is_closed=true
   Output: {"macro_actions":["cover_area"], "confidence":0.97}
\end{lstlisting}
  \end{minipage}
\end{figure*}

\section{Elaboration Study: Zero-shot vs One-shot Prompting}
\label{app:elab}

We assess the sensitivity of segment-level macro-action selection to in-context prompting during inference. The network weights are fixed; only the content of the language channel \(L\) differs. We compare: (1) Zero-shot ($L_0^{\mathrm{ZS}}$): the default prompt of Sec.~\ref{app:prompt} without the [EXAMPLES] block; (2) One-shot ($L_0^{\mathrm{1S}}$): the same prompt with a single in-context example chosen to match the current segment type. For path segments we provide one example aligned to the segment’s coarse turn class (straight, \(\pm45^{\circ}\), \(\pm90^{\circ}\)). For area segments we provide one coverage example. The one-shot exemplar is drawn from a held-out seed pool and never reuses the test scene. All other inputs remain unchanged: the system processes one segment \(s_k\) at a time with \((I,S,L)\) and produces a single macro-action token \(a'_k\in\mathcal{A}_{\text{disc}}\) per \(s_k\). As shown in Table \ref{tab:elab_prompt}, we report segment-level macro-action accuracy against annotation ground truth \(\mathcal{R}_{\text{ann}}\), along with downstream SSSR, FTCR, and FTSPAR aggregated on the HouseholdSketch validation split.

\begin{table}[t]
\centering
\caption{Zero-shot vs. one-shot prompting on HouseholdSketch. Values are percentages. One-shot adds a single segment-matched exemplar to \(L_0\).}
\label{tab:elab_prompt}
\begin{tabular}{lcccc}
\hline
Setting & Macro Acc. & SSSR & FTCR & FTSPAR \\
\hline
\(L_0^{\mathrm{ZS}}\) (zero-shot) & 88.4 & 84.0 & 74.2 & 62.0 \\
\(L_0^{\mathrm{1S}}\) (one-shot)  & \textbf{89.8} & \textbf{85.1} & \textbf{75.2} & \textbf{63.0} \\
\hline
\end{tabular}
\end{table}

To examine length sensitivity we further break out SSSR, FTCR, and FTSPAR by task complexity category. The results are shown in Table \ref{tab:elab_prompt_len}.

\begin{table}[t]
\centering
\caption{Length-conditioned comparison for zero-shot and one-shot prompting.}
\label{tab:elab_prompt_len}
\begin{tabular}{lcccc}
\hline
\multicolumn{5}{c}{SSSR (\%)}\\
\hline
 & Short & Medium & Long & Overall \\
\hline
\(L_0^{\mathrm{ZS}}\) & 84.7 & 84.2 & 83.8 & 84.0 \\
\(L_0^{\mathrm{1S}}\) & \textbf{85.5} & \textbf{85.0} & \textbf{84.8} & \textbf{85.1} \\
\hline
\multicolumn{5}{c}{FTCR (\%)}\\
\hline
 & Short & Medium & Long & Overall \\
\hline
\(L_0^{\mathrm{ZS}}\) & 74.8 & 60.0 & 46.4 & 74.2 \\
\(L_0^{\mathrm{1S}}\) & \textbf{75.6} & \textbf{60.8} & \textbf{47.3} & \textbf{75.2} \\
\hline
\multicolumn{5}{c}{FTSPAR (\%)}\\
\hline
 & Short & Medium & Long & Overall \\
\hline
\(L_0^{\mathrm{ZS}}\) & 62.7 & 47.9 & 33.3 & 62.0 \\
\(L_0^{\mathrm{1S}}\) & \textbf{63.4} & \textbf{48.6} & \textbf{34.1} & \textbf{63.0} \\
\hline
\end{tabular}
\end{table}

One-shot prompting yields consistent but modest gains across all metrics. Improvements are most visible in macro-action accuracy and per-segment SSSR, which then propagate to small increases in FTCR and FTSPAR. Gains are slightly larger for Long tasks, where minor disambiguation at turn boundaries reduces compounding classification errors across \(\mathcal{S}_{\text{seq}}\). The absolute differences remain limited because the dominant information comes from sketch geometry \(S\) and the photograph \(I\); the language channel acts as a structured prior that nudges decisions in borderline cases without altering the discrete vocabulary \(\mathcal{A}_{\text{disc}}\) or the control parameters used by \(g_{\text{translate}}\). For reproducibility, all results were obtained with the same frozen visual and language encoders, identical fusion and policy weights, and identical preprocessing; only the presence or absence of a single in-context exemplar in \(L_0\) changed.

\section{Ablation Study on Live Perception}
\label{app:pt_ablation}

This section complements Section~VI.B (in the main paper) by quantifying the effect of optional live perception \(P_t\) when injected through the image channel at inference time. As described in Section~VI.B, when enabled we form \((I,S,L,P_t)\) and encode \(P_t\) with the same frozen visual backbone \(\phi_V\) used for \(I\). All visual frames are resized to \(224\times224\) and normalized with ImageNet statistics to match \(\phi_V\). In this release \(P_t\) is RGB from the robot camera; depth and LiDAR remain within the platform safety and navigation controllers and are not fed to \(\psi_{\text{fuse}}\) for the default model. Because \(I\) (third person) and \(P_t\) (egocentric) come from different viewpoints, we do not perform explicit warping; sketches are grounded in \(I\), while \(P_t\) contributes egocentric evidence for reactive checks. Late fusion follows
\[
F^{\text{fused}}_{k}=\mathrm{MLP}_{\text{fuse}}\!\big([\,f^{S}_{k};\, f^{V}_{\text{att}}(I);\, f^{V}_{\text{cls}}(I);\,
f^{L};\, \eta_t\, f^{V,\text{live}}_{\text{cls}}(P_t)\,]\big),
\]
with a binary gate \(\eta_t\). In our implementation \(\eta_t=1\) during obstacle-handling events (for example \texttt{check\_under} and post-detection clearance assessment) and \(\eta_t=0\) otherwise, which leaves \((I,S,L)\) as the dominant drivers of macro-action selection.

We evaluated four inference variants without retraining the model: (A) baseline \((I,S,L)\) only, (B) late-fusion gating of \(P_t\) as above, (C) naive early fusion that concatenates \(P_t\) patch tokens with \(I\) tokens before attention, and (D) depth-injected \(P_t\) where the depth map is linearly scaled to \([0,255]\), replicated to 3 channels, and passed through \(\phi_V\). We report results on a 1{,}200-trial obstacle subset from HouseholdSketch in which at least one segment required an under-obstacle clearance check, stratified across scenes and sketch lengths. We also summarize averages over the full validation set. For the obstacle subset we additionally report the Under-Obstacle Maneuver Success (UOMS), defined as the fraction of obstacle encounters that are correctly classified as traversable or not and handled without manual intervention.

\begin{table}[t]
\centering
\scriptsize
\caption{Live perception ablation on the obstacle subset (1{,}200 trials). Percentages are averages across scenes and sketch-length strata.}
\label{tab:pt_obstacle_subset}
\begin{tabular}{lcccc}
\hline
Setting & SSSR & SSSPAR & FTCR & UOMS \\
\hline
(A) \((I,S,L)\) & 84.1 & 76.7 & 58.2 & 62.5 \\
(B) \((I,S,L)+P_t\) late-fusion gated & \textbf{84.3} & \textbf{77.5} & \textbf{64.9} & \textbf{78.4} \\
(C) \((I,S,L)+P_t\) naive early fusion & 83.9 & 75.9 & 60.1 & 70.2 \\
(D) \((I,S,L)+P_t^{\text{depth}}\) (replicated) & 84.0 & 76.1 & 59.0 & 67.3 \\
\hline
\end{tabular}
\end{table}

Table~\ref{tab:pt_obstacle_subset} shows that gated late fusion of \(P_t\) yields a clear improvement on the obstacle subset: FTCR increases by \(+6.7\) points relative to the baseline, and UOMS increases by \(+15.9\) points, confirming that live egocentric evidence is most valuable when deciding whether to proceed under furniture. Changes to SSSR and SSSPAR are small, which is expected because segment-wise action selection remains dominated by \((I,S,L)\) and because continuous smoothing occurs in the platform controllers. Naive early fusion underperforms the gated design, suggesting that mixing egocentric and third-person frames at the token level without viewpoint reasoning dilutes attention. Injecting depth into \(\psi_{\text{fuse}}\) without modality-specific training does not surpass RGB \(P_t\); this aligns with our internal observation that depth benefits require tailored pretraining or adapters.

Over the full validation set (all scenes, all sketch lengths), enabling gated \(P_t\) produces negligible change in SSSR (84.4 to 84.6) and a modest FTCR increase of \(+0.9\) points, with the gain concentrated in scenes that frequently trigger obstacle checks (for example dining areas with tables and chairs). These results justify our default of using \(P_t\) selectively through \(\eta_t\): \((I,S,L)\) provides a strong global prior for intent grounding, while live perception is activated precisely where it adds value.

\end{document}